\documentclass[3p,a4paper]{elsarticle}
\usepackage[utf8]{inputenc}
\usepackage{subfigure}

\title{Neural Tangent Kernel Analysis to Probe Convergence in Physics-informed Neural Solvers: PIKANs vs.~PINNs
}

\author[myUaddress]{Salah A. Faroughi \corref{mycorrespondingauthor}}
\author[myUaddress]{Farinaz Mostajeran}

\cortext[mycorrespondingauthor]{Corresponding author: salah.faroughi@utah.edu}

\address[myUaddress]{Energy \& Intelligence Lab, Department of Chemical Engineering, University of Utah, Salt Lake City, Utah  84112, USA
}

\date{\today}

\usepackage{mwe}

\usepackage{setspace} 
\usepackage{tabularx}
\usepackage{lineno,hyperref}
\usepackage{amsmath}
\usepackage{bm}
\usepackage{amssymb}
\usepackage[]{amsmath}
\usepackage{upgreek}
\usepackage{float}

\let\today\relax
\makeatletter
\def\ps@pprintTitle{%
    \let\@oddhead\@empty
    \let\@evenhead\@empty
    \def\@oddfoot{\footnotesize\itshape
         {Submitted preprint — June 2025} \hfill\today}%
    \let\@evenfoot\@oddfoot
    }
\makeatother

\usepackage{pgfplots}
\pgfplotsset{compat=1.5}
\usepgfplotslibrary{units}
\usetikzlibrary{spy}
\usetikzlibrary{pgfplots.groupplots}

\usepackage{epstopdf}
\usepackage{units}
\usepackage{lscape}
\usepackage{booktabs}
\usepackage{gensymb}
\usepackage{multirow}

\usepackage{graphicx}
\usepackage{caption}
\usepackage{subcaption}

\usepackage{booktabs}
\usepackage{multirow}
\usepackage{caption}
\usepackage{subcaption}

\usepackage{soul,color}
\soulregister\cite7
\soulregister\ref7
\soulregister\pageref7

\usepackage[flushleft]{threeparttable}
\usepackage{tikz}
\usepackage{placeins}
\usepackage{color}
\usepackage{hyperref}

\usepackage{stackengine}

\usepackage{array}
\usepackage{tabularx}
\usepackage{pdfpages}
\usepackage{natbib}
\biboptions{numbers,sort,compress} 
\usepackage{textcomp, gensymb}
\usepackage{booktabs} 

\newtheorem{theorem}{Theorem}

\newtheorem{lemma}{Lemma}

\newtheorem{remark}{Remark}

\newtheorem{open problem}[theorem]{Open Problem}
{\par\noindent\textbf{Proof.}\ }%
{\hfill$\square$\par}

\begin{document}

\begin{abstract}

Physics-informed Kolmogorov--Arnold Networks (PIKANs), and in particular their Chebyshev-based variants (cPIKANs), have recently emerged as promising models for solving partial differential equations (PDEs). However, their training dynamics and convergence behavior remain largely unexplored both theoretically and numerically. In this work, we aim to advance the theoretical understanding of cPIKANs by analyzing them using Neural Tangent Kernel (NTK) theory. Our objective is to discern the evolution of kernel structure throughout gradient-based training and its subsequent impact on learning efficiency. We first derive the NTK of standard cKANs in a supervised setting, and then extend the analysis to the physics-informed context. We analyze the spectral properties of NTK matrices, specifically their eigenvalue distributions and spectral bias, for four representative PDEs: the steady-state Helmholtz equation, transient diffusion and  Allen–Cahn equations, and forced vibrations governed by the Euler–Bernoulli beam equation.
We also conduct an investigation into the impact of various optimization strategies, e.g., first-order, second-order, and hybrid approaches, on the evolution of the NTK and the resulting learning dynamics. Results indicate a tractable behavior for NTK in the context of cPIKANs, which exposes learning dynamics that standard physics-informed neural networks (PINNs) cannot capture.  Spectral trends also reveal when domain decomposition improves training, directly linking kernel behavior to convergence rates under different setups. To the best of our knowledge, this is the first systematic NTK study of cPIKANs, providing theoretical insight that clarifies and predicts their empirical performance.
 
\end{abstract}


\maketitle

\section{Introduction}\label{sec:Intro}

Modeling systems characterized by diverse phenomena occurring across different scales is a computational challenge. Some systems possess dynamics spanning nanometers to kilometers spatially, and nanoseconds to years temporally. Such problems arise in a range of real-world applications, including nuclear reactor design and safety \cite{gaston2015physics}, catalysts and functional materials \cite{cullen2021new,zhou2021hybrid}, combustion dynamics \cite{xi2024engineering,lee2024large}, synthetic biology and bioengineering \cite{fotiadis2023multiscale}, battery systems \cite{yu2023lithium}, and power grid infrastructures \cite{gao2022multiscale}. More complex examples are found in multiphase reacting flows within porous media \cite{pawar2024geo} and Earth systems modeling under uncertainty \cite{mahjour2024selection,mahjour2025select},  both of which can exhibit dynamics over vastly extended spatial and temporal scales.  In order to investigate such systems, multifaceted mathematics are needed to capture the system in its entirety  while  integrating all components \cite{heinlein2025multifidelity, zhou2025multilevel, van2025multiscale}. This approach  typically leads to a set of coupled partial or ordinary differential equations (PDEs and ODEs), cross-scale closure models, stochastic processes, and uncertainty analytics. When modeling these systems, especially in an inverse context where extensive optimization is required to identify targeted properties or infer unknown characteristics \cite{faroughi2024physics,kim2024review}, rapid convergence and high accuracy become essential. Classical numerical methods, such as finite difference, finite element, and finite volume schemes, are effective, but their computational cost scales poorly with domain size, $\Omega \times [0, T]$, due to stability restrictions (e.g., CFL conditions) and the need for fine resolution to capture localized phenomena \cite{oladayo2025stability, orucc2025numerical, weiss2024effects}.

To overcome the computational limitations of classical solvers, physics-informed neural networks (PINNs) have been introduced as a mesh-free alternative that incorporates the governing PDEs directly into the loss function of a deep neural network \cite{raissi2019physics, raissi2018hidden, karniadakis2021physics, mostajeran2022deepbhcp}. Over the past few years, PINNs have demonstrated remarkable success across a wide range of scientific and engineering problems, showcasing their potential as general-purpose solvers for forward and inverse modeling. 
Notable applications include modeling turbulent fluid flows \cite{hanrahan2023studying, jang2024physics, yazdani2024data, gafoor2025physics} and viscoelastic fluid flows \cite{mahmoudabadbozchelou2022nn, thakur2024viscoelasticnet}, solving cardiovascular flow problems \cite{kharazmi2021hp, arzani2021data, zhang2023physics},  simulating seismic wave propagation \cite{rasht2022physics, zou2025accelerating, zou2024seismic}, and identifying material parameters in solid mechanics \cite{jagtap2022physics, wu2023effective, hu2024physics, teloli2025physics}. 
These studies highlight the strength of PINNs in leveraging limited data and physical constraints to infer complex dynamics, especially in scenarios where traditional solvers are either too costly or inapplicable.  
Despite their theoretical appeal and success on benchmark problems, PINNs often perform poorly on large domains, particularly when the solution exhibits high-frequency oscillations, sharp interfaces, or multiscale structures \cite{mishra2023estimates, krishnapriyan2021characterizing, wang2021eigenvector}. Several studies have investigated the underlying failure modes of PINNs and attempted to characterize their learning dynamics \cite{yu2022gradient, li2023physics, basir2022critical, shi2024physics}. These limitations have motivated the development of alternative neural architectures designed to improve expressivity, training stability, and convergence in the presence of complex physical behavior.


To overcome the limitations of MLP-based PINNs, recent advances have introduced alternative architectures grounded in the Kolmogorov--Arnold representation theorem 
\cite{kolmogorov1957representations, Arnold1958}. Among them, Kolmogorov–Arnold Networks (KANs) \cite{liu2024kan, Kolmogorov1957, Arnold1957} replace fixed nonlinear activations with learnable univariate functions, often implemented using splines or low-order polynomials. 
This formulation provides finer control over spectral bias and improves approximation of localized or high-frequency features \cite{shukla2024comprehensive}.
When integrated with the physics-informed training paradigm, the resulting physics-informed KANs (PIKANs) offer enhanced expressivity while remaining grounded in physical laws \cite{shukla2024comprehensive, mostajeran2024epi, mostajeran2025scaled}. In particular, the Chebyshev-based variant, cPIKAN, leverages orthogonal polynomial bases to improve stability and interpretability in  learning tasks \cite{ss2024chebyshev, guo2024physics}. However, the use of Chebyshev polynomials necessitates input normalization to the $[-1,1]$ interval for numerical stability, especially in large or multi-scale domains \cite{ss2024chebyshev}. To this end, Scaled-cPIKAN \cite{mostajeran2025scaled} introduces a scaling strategy for spatial variables and residual losses that enhances convergence and accuracy, as verified across multiple benchmark problems. By aligning the architectural flexibility of KANs with domain-informed normalization, this approach effectively addresses the performance bottlenecks of vanilla PINNs in extended spatial domains. 
Also, in \cite{mostajeran2025scaled}, the authors further investigate the theoretical necessity of variable scaling by analyzing the Neural Tangent Kernel (NTK) structure \cite{jacot2018neural}, leveraging the mathematical simplicity of Chebyshev polynomials to establish a clear justification for their scaling strategy.
These developments set the stage for a deeper theoretical understanding of training dynamics in cPIKANs, particularly through the lens of the NTK analysis, which we want to explore in this paper.

The NTK was introduced by Jacot et al. \cite{jacot2018neural} as a theoretical framework to analyze the training dynamics of infinitely-wide neural networks under gradient descent. They showed that, in the infinite-width limit, neural networks behave like kernel methods, where the NTK converges to a deterministic, constant kernel during training. This allows the study of learning dynamics in function space, rather than parameter space, and connects convergence properties to the eigenstructure of the NTK. Their work also demonstrated that the speed of convergence depends on the eigenvalues of the NTK, with faster convergence along directions corresponding to larger eigenvalues. Building on this, \cite{seleznova2022analyzing} investigated the validity of NTK theory for finite-width networks. Their results revealed that the NTK approximation often fails for deep or improperly initialized networks, where gradients may vanish or explode. They further showed that NTK behavior depends critically on the depth-to-width ratio and the initialization regime \cite{seleznova2022neural}, whether the network operates in the ordered phase, chaotic phase, or at the edge of chaos. In particular, they proved that the variability of the NTK grows exponentially with depth in the chaotic and the edge of chaos regimes, undermining the assumption of a constant NTK during training.
Inspired by these findings, Wang et al. \cite{wang2022and} applied NTK analysis to PINNs. They derived the NTK for PINNs and showed that, under infinite-width assumptions, it converges to a deterministic kernel that remains nearly constant during training. However, they identified a critical limitation: an imbalance in convergence rates across different loss components, driven by spectral bias. To mitigate this issue, they proposed an adaptive gradient descent algorithm that uses NTK eigenvalues to balance learning dynamics. Saadat et al. \cite{saadat2022neural} extended these ideas to PINNs for linear advection-diffusion equations and demonstrated that variations in advection speed or diffusion coefficient can cause training failures. Their work highlighted that PINNs often struggle to learn initial or boundary conditions when PDE parameters dominate. They also suggested strategies such as adaptive loss weighting and periodic activation functions to address these challenges.

In this work, we aim to advance the theoretical understanding of cPIKANs by studying their training behavior through the NTK framework. We focus on how the use of Chebyshev polynomial bases, combined with input normalization strategies, shapes the NTK eigenstructure, and how these factors affect the learning efficiency and generalization of cPIKANs on large or multi-scale domains. Understanding these relationships is essential for designing architectures that balance expressiveness, stability, and computational efficiency when modeling complex physical systems with sharp interfaces or high-frequency features.
Our analysis begins by deriving explicit expressions for the NTK of cKANs in the standard, non-physics-informed setting. We study how the choice of basis functions and network parameters influences the NTK structure for finite-width networks. We then extend the analysis to the physics-informed case, deriving the NTK for cPIKANs, where PDE residuals are included in the loss. We provide a detailed breakdown of the NTK into components that reflect the effects of physics-informed terms.
We support our theoretical findings with experiments on various PDE examples. First, we compare the NTK behavior of cKANs with standard MLPs. Then, we study cPIKANs trained with different optimization algorithms such as Adam and L-BFGS, and compare the impact of Chebyshev and B-spline basis functions on the NTK. We also analyze how the NTK structure varies across different subdomains within a single problem.
Overall, this work bridges the gap between empirical observations and theoretical understanding in cPIKANs. The insights provided here can guide the development of more robust and efficient physics-informed neural architectures for complex physical modeling tasks.

\section{Preliminary}

\subsection{Kolmogorov-Arnold Networks (KANs)}
\label{sec:kolmogorov_arnold_networks}
Kolmogorov’s superposition theorem, introduced in 1957 \cite{kolmogorov1957representations} and refined by Arnold \cite{Arnold1958,arnold1959representation,arnold2009functions}, states that any continuous multivariate function can be expressed as a combination of continuous single-variable functions.  This result was extended by Lorentz \cite{lorentz1966approximation,lorentz1996constructive}, Sprecher \cite{sprecher1965structure,sprecher1972improvement}, and Friedman \cite{fridman1967improvement}, though early constructions had issues such as the non-smoothness of the inner functions \cite{hecht1987kolmogorov}. Later,  K\r{u}rkov{\'a}  \cite{kuurkova1991kolmogorov,kuurkova1992kolmogorov} addressed these concerns using sigmoidal approximations, while Sprecher and K\"{o}ppen proposed numerical and structural improvements \cite{sprecher1996numerical,koppen2002training}, culminating in a constructive proof by Braun and Griebel \cite{braun2009constructive}. This Kolmogorov-Arnold Representation Theorem forms a theoretical basis for universal function approximation \cite{cybenko1989approximation, hornik1991approximation, barron1993universal} and continues to influence the design of neural architectures that use compositions (depth) and summations (width) of simple univariate functions to approximate complex mappings \cite{telgarsky2016benefits, lu2017expressive, schmidt2020nonparametric, poggio2017and}.

Early implementations of the Kolmogorov-Arnold representation in neural networks followed the original depth-2, width-$(2n + 1)$ structure \cite{sprecher2002space, koppen2002training}, but faced practical limitations due to non-smooth inner functions and inefficient training. Recent advances \cite{fakhoury2022exsplinet, he2023optimal} have renewed interest in this approach by incorporating gradient-based optimization and emphasizing interpretability, particularly in scientific domains. These developments have led to the emergence of Kolmogorov-Arnold Networks (KANs) \cite{Kolmogorov1957, Arnold1957, Sun2021, Chang2022}, which explicitly implement the decomposition,
\begin{equation}
    f(x_1, \ldots, x_d) = \sum_{k=0}^{2d} \varphi_k \left( \sum_{j=1}^{d} \psi_{k,j}(x_j) \right),
\end{equation}
using trainable univariate functions $\psi_{k,j}$ and $\varphi_k$ within neural architectures. While this formulation corresponds to a depth-2 network, recent work \cite{liu2024kan} extends KANs to multi-layer architectures by composing univariate functions, for example,
\begin{equation}
    \phi_k = \phi_k^{(L)} \circ \cdots \circ \phi_k^{(1)},
\end{equation}
which improves both smoothness and representational capacity. In this setting, a KAN with $L$ layers can be expressed as a composition of layer-wise basis functions,
\begin{equation}
    f(x) = (\Phi_{L-1} \circ \Phi_{L-2} \circ \cdots \circ \Phi_0)(x),
\end{equation}
where each $\Phi_\ell: \mathbb{R}^{d_\ell} \to \mathbb{R}^{d_{\ell+1}}$ consists of univariate transformations applied element-wise, and the outputs are recursively computed as $x_{\ell+1} = \Phi_\ell(x_\ell)$. This generalized formulation retains the theoretical foundation of Kolmogorov’s superposition while harnessing the expressive power of deep architectures to achieve greater flexibility and efficiency.

Kolmogorov-Arnold Networks rely on carefully chosen univariate basis functions for approximation.  B-splines  are a common choice \cite{}, offering local control through piecewise polynomial functions defined on a grid, 
\begin{equation}
\phi(x) = w_b\,b(x) + w_s \sum_i c_i B_i(x), \quad b(x) = \frac{x}{1 + e^{-x}},
\end{equation}
where $B_i(x)$ are spline basis functions defined by the grid size $g$ and polynomial order $k$, with trainable parameters $\boldsymbol{\theta} = \{c_i, w_b, w_s\}$. While effective for structured data, spline-based KANs scale poorly with increasing model complexity, requiring $O(N_l N_n^2 (k + g))$ parameters. To enhance the modeling of fine-grained and hierarchical features, researchers have developed wavelet-based KANs (Wav-KAN) \cite{bozorgasl2405wav}. These leverage multiscale wavelet expansions to efficiently capture localized patterns, with the number of trainable parameters scaling as $O(3N_l N_n^2)$. For problems requiring smooth interpolation, radial basis function (RBF) networks \cite{li2024kolmogorov} provide an effective alternative, maintaining the same parameter complexity while offering superior approximation properties for continuous functions.

An increasingly favored approach for KAN design involves the use of Chebyshev polynomials \cite{ss2024chebyshev} as a globally orthogonal basis \cite{rivlin2020chebyshev, schmidt2021kolmogorov}, offering strong theoretical guarantees in approximation and efficient parameter usage. In the Chebyshev-KAN (cKAN) framework, the nonlinear mapping is expressed via a polynomial expansion,
\begin{equation}
\phi(x) = \sum_{n=0}^k c_n\, T_n(x),
\end{equation}
where $T_n(x)$ denotes the $n$th Chebyshev polynomial of the first kind, recursively defined as,
\begin{equation}
T_0(x) = 1,\quad T_1(x) = x,\quad T_n(x) = 2\: x\: T_{n-1}(x) - T_{n-2}(x).
\end{equation}
This orthogonal basis enables stable and efficient function approximation, especially for smooth or oscillatory inputs, while reducing the parameter count to $|\boldsymbol{\theta}| \sim O(N_l N_n^2 k)$. For faster evaluation, a trigonometric form $T_n(x) = \cos(n\,\arccos(x))$ is sometimes used, though care must be taken near the domain boundaries $x \in [-1,1]$ to avoid numerical instability \cite{mostajeran2025scaled}. To ensure numerical stability and input compatibility, inputs are normalized using $\tanh$ activations across layers \cite{guo2024physics, hu2024tackling, ss2024chebyshev}, leading to the composite architecture,
\begin{equation}
f_{\text{cKAN}}(\boldsymbol{x}) = (\boldsymbol{\Phi}_L \circ \tanh \circ \cdots \circ \tanh \circ \boldsymbol{\Phi}_1 \circ \tanh)(\boldsymbol{x}).
\end{equation}
The Chebyshev basis thus offers a computationally efficient and theoretically grounded method for representing high-order nonlinearities in KANs, particularly beneficial in applications involving high-frequency content or rapidly varying target functions.

\subsection{Neural Tangent Kernel (NTK)}

The Neural Tangent Kernel, introduced in the seminal work by Jacot et al. \cite{jacot2018neural}, provides a powerful theoretical framework for understanding the training dynamics of neural networks. NTK theory studies fully-connected neural networks in the infinite-width limit, where the number of neurons in each hidden layer tends to infinity \cite{jacot2018neural, arora2019exact}. In this regime, it has been shown that the network's behavior during training can be approximated by a linear model derived from the first-order Taylor expansion around its initialization \cite{jacot2018neural, arora2019exact, lee2019wide}. Under such conditions and a specific parameter initialization, the NTK remains constant throughout training, rendering the training process equivalent to deterministic kernel regression. This insight reveals that training an over-parameterized neural network with gradient descent can be viewed through kernel methods, providing a clearer understanding of why such networks often generalize well despite their complexity.

In \cite{mostajeran2025scaled}, the authors showed that scaling spatial variables is crucial for stable and efficient learning in the cKAN framework using NTK analysis. Their study revealed that the spectral properties of the NTK matrix strongly influence the training dynamics and highlighted the role of proper scaling in achieving robust model performance. 
Let us consider the problem of minimizing the squared loss function,
\begin{equation}\label{Eq.SmoothLoss}
\mathcal{L}(\boldsymbol{\theta}) = \frac{1}{2N}\sum_{i=1}^{N}(y_i - f(\boldsymbol{x}_i; \boldsymbol{\theta}))^2,
\end{equation}
over a dataset $\mathcal{D} = \{(\boldsymbol{x}_i, y_i) \}_{i=1}^{N} \subset [-1,1]^{d_{\text{in}}} \times \mathbb{R}$, where $f$ denotes the output of a cKAN parameterized by $\boldsymbol{\theta}$. The continuous version of gradient descent, known as gradient flow, is given by,
\begin{equation}
\frac{d \boldsymbol{\theta}(\tau)}{d \tau} = - \nabla_{\boldsymbol{\theta}} l(\boldsymbol{\theta}(\tau)) = 
- \sum_{i=1}^{N} \left(y_i - f(\boldsymbol{x}_i; \boldsymbol{\theta})\right)  \nabla_{\boldsymbol{\theta}} f(\boldsymbol{x}_i; \boldsymbol{\theta}).
  \end{equation}
Throughout this work, $\tau$ denotes the gradient flow time, a continuous-time variable that represents the evolution of the network parameters under gradient descent. For simplicity, we refer to it as training time.
Using the chain rule, the time evolution of the network output can be expressed as, 
\begin{equation}
\frac{d f(\boldsymbol{x}_i; \boldsymbol{\theta})}{d \tau} = \left[ \frac{d f(\boldsymbol{x}_i; \boldsymbol{\theta})}{d \boldsymbol{\theta}}\right] \cdot \left[\frac{d \boldsymbol{\theta}(\tau)}{d \tau}\right] = 
- \sum_{i=1}^{N} \left(y_i - f(\boldsymbol{x}_i; \boldsymbol{\theta})\right) \left[\nabla_{\boldsymbol{\theta}} f(\boldsymbol{x}_i; \boldsymbol{\theta})\right]^T  \left[\nabla_{\boldsymbol{\theta}} f(\boldsymbol{x}_i; \boldsymbol{\theta})\right].
  \end{equation}
The Neural Tangent Kernel matrix associated with the cKAN model is defined as,
\begin{equation}\label{Eq:K_NTK_ij}
(\boldsymbol{K}_{\text{ntk}}(\tau))_{i,j} = \left\langle \frac{\partial f(\boldsymbol{x}_i; \boldsymbol{\theta}(\tau))}{\partial \boldsymbol{\theta}}, \frac{\partial f(\boldsymbol{x}_j; \boldsymbol{\theta}(\tau))}{\partial \boldsymbol{\theta}} \right\rangle.
\end{equation}
$\boldsymbol{K}_{\text{ntk}}(\tau)$ is a positive semi-definite block matrix that captures the sensitivity of each output component with respect to changes in model parameters.
Finally, the evolution of the network predictions $\boldsymbol{v}(\tau) = \left(f(\boldsymbol{x}_1; \boldsymbol{\theta}(\tau)), \ldots, f(\boldsymbol{x}_N; \boldsymbol{\theta}(\tau))\right) \in \mathbb{R}^N$ follows the differential equation, 
\begin{equation} \label{Eq.NTK1}
\frac{d \boldsymbol{v}(\tau)}{d \tau} = - \boldsymbol{K}_{\text{ntk}}(\tau) \cdot (\boldsymbol{v}(\tau) - \boldsymbol{y}),
\end{equation}
where $\boldsymbol{y} = (y_1, \ldots, y_N)$ represents the target outputs. This formulation reveals how the NTK governs the dynamics of training in the cKAN framework. Assume that the output of the cKAN model can be represented as a Chebyshev polynomial expansion,
\begin{equation}\label{Eq.Assumption1}
f(x; \boldsymbol{\theta}) = \sum_{n=0}^{k} c_n T_n(x),
\end{equation}
where $k$ is the number of terms in the expansion, and $\boldsymbol{\theta} = \{c_n\}_{n=0}^{k}$ denotes the corresponding coefficients. 
Since the partial derivative of the cKAN output in Eq.~\eqref{Eq.Assumption1} with respect to each coefficient $c_n$ is simply the corresponding Chebyshev polynomial,
\begin{equation}
\frac{\partial f}{\partial c_n} = T_n(x) \quad \text{for all } n = 0, \ldots, k,
\end{equation}
the Neural Tangent Kernel takes the form,
\begin{equation}\label{Eq.KcKAN}
(\boldsymbol{K}_{\text{ntk}}(\tau))_{i,j} = \sum_{n=0}^{k} T_n(x_i)\: T_n(x_j).
\end{equation}
This shows that the NTK matrix is independent of the parameters $c_n$ and remains constant throughout training,
\begin{equation}
\boldsymbol{K}_{\text{ntk}}^{*} = \boldsymbol{K}_{\text{ntk}}(\tau) \quad \text{for all } \tau.
\end{equation}
Therefore, under the assumption in Eq.~\eqref{Eq.Assumption1}, the NTK does not evolve during training \cite{mostajeran2025scaled}.

\section{Theoretical Development}

This section  develops the theoretical framework needed to analyze the NTK associated with the cPIKAN model. We recently studied the NTK of cKAN using a non-nested approximation that simplifies the treatment of input scaling and enables analytical analysis \cite{mostajeran2025scaled}.  While this approach provides useful insight into the influence of the Chebyshev basis on the NTK structure, it does not capture the full complexity of the nested cKAN architecture. To address this gap, we first investigate the NTK associated with cKANs in a nested  data-driven setting; a  theoretical analysis that lays the foundation for the second part, where we extend the analysis to  cPIKANs  by incorporating physics-informed constraints into the NTK framework.

\subsection{Step I: NTK for cKANs}

This section analyzes the behavior of $\boldsymbol{K}_{\text{ntk}}(\tau)$ under the use of nested approximation.
We consider a cKAN with a single hidden layer consisting of $N$ neurons, input dimensionality $d$, Chebyshev polynomial degree $k$, and a scalar output. For each normalized input coordinate $\tilde{x}_i = \tanh(x_i)$, $i = 1, \dots, d$, the Chebyshev polynomials of the first kind up to degree $k$ are computed recursively as,
\begin{equation}
T_0(\tilde{x}_i) = 1, \quad
T_1(\tilde{x}_i) = \tilde{x}_i,\quad
T_n(\tilde{x}_i) = 2\:\tilde{x}_i
\: T_{n-1}(\tilde{x}_i) - T_{n-2}(\tilde{x}_i), \quad n=2, \dots , k,
\end{equation}
and the output of each hidden neuron $j$ is given by,
\begin{equation}
h_j(\boldsymbol{x}) = \sum_{i=1}^{d} \sum_{n=0}^{k} w_{i,j,n}^{(1)} \cdot T_n(\tilde{x}_i),\quad j = 1, \dots, N,
\end{equation}
where $w_{i,j,n}^{(1)}$ are trainable coefficients associated with input coordinate $i$, neuron $j$, and polynomial degree $n$.
The final output $f$ is computed as a linear combination of the transformed hidden neurons,
\begin{equation}
f(\boldsymbol{x}; \boldsymbol{\theta}) = \sum_{j=1}^{N} \sum_{n=0}^{k} w_{j,1,n}^{(2)} \cdot T_n(\tanh(h_j(\boldsymbol{x}))),
\end{equation}
where $w_{j,1,n}^{(2)}$ are trainable coefficients associated with hidden neuron $j$.
The full set of trainable parameters is denoted as $\boldsymbol{\theta} = \{w_{i,j,n}^{(1)},  w_{j,1,n}^{(2)}\}$. We can also show that,
\begin{equation}
    \frac{\partial h_j(\boldsymbol{x})}{\partial w_{i,j,n}^{(1)}} = T_n(\tilde{x}_i), 
    \quad
    \frac{\partial f(\boldsymbol{x}; \boldsymbol{\theta})}{\partial h_j} = \bigl(1 - \tanh^2(h_j(\boldsymbol{x}))\bigr) \cdot 
\sum_{n=0}^k w_{j,1,n}^{(2)} \, T_n'\bigl(\tanh(h_j(\boldsymbol{x}))\bigr),
\end{equation}
where \( T_n'(\cdot) \) denotes the derivative of the Chebyshev polynomial \( T_n \) with respect to its argument.
Therefore, the derivative of the output with respect to the first-layer coefficients is,
\begin{equation}\label{Eq:firstdiff}
\frac{\partial f(\boldsymbol{x}; \boldsymbol{\theta})}{\partial w_{i,j,n}^{(1)}} 
= T_n(\tilde{x}_i) \cdot \bigl(1 - \tanh^2(h_j(\boldsymbol{x}))\bigr) \cdot 
\sum_{m=0}^k w_{j,1,m}^{(2)} \, T_m'\bigl(\tanh(h_j(\boldsymbol{x}))\bigr).
\end{equation}
and,  the derivative of the output with respect to the second-layer coefficients is,
\begin{equation}
    \frac{\partial f(\boldsymbol{x}; \boldsymbol{\theta})}{\partial w_{j,1,n}^{(2)}} = T_n\bigl(\tanh(h_j(\boldsymbol{x}))\bigr).
\end{equation}

Given that the coefficients $w_{i,j,n}^{(1)}$ and $w_{j,1,m}^{(2)}$ are independently and identically distributed as standard normal random variables,
\begin{equation}
w_{i,j,n}^{(1)},\; w_{j,1,m}^{(2)} \overset{\text{iid}}{\sim} \mathcal{N}(0,1),
\end{equation}
and assuming that the input values $\boldsymbol{x} = (x_1, \dots , x_d)$ are fixed, the hidden pre-activation for neuron $j$ is given by,
\begin{equation}
h_j(\boldsymbol{x}) = \sum_{i=1}^d \sum_{n=0}^k w_{i,j,n}^{(1)}\,T_n(\tanh(x_i)) \sim \mathcal{N}(0,\;\sigma_{h_j}^2),
\end{equation}
where the variance $\sigma_{h_j}^2$ is determined by,
\begin{equation}\label{Eq:VarianceH}
\sigma_{h_j}^2 = \sum_{i=1}^d \sum_{n=0}^k \left[T_n(\tanh(x_i))\right]^2.
\end{equation}

Since $h_j$ is a Gaussian random variable, the transformed value $z_j = \tanh(h_j)$ is a smooth and bounded function of a Gaussian variable. While $z_j$ is no longer Gaussian, its distribution remains symmetric, centered at zero, and unimodal.
The network output $f$ can thus be computed as a linear combination of the second-layer coefficients $w_{j,1,n}^{(2)}$, weighted by the Chebyshev polynomials evaluated at $z_j$,
\begin{equation}
f(\boldsymbol{x}) = \sum_{j=1}^{N} \sum_{n=0}^{k} w_{j,1,n}^{(2)} \cdot T_n(z_j(\boldsymbol{x})),
\end{equation}
where $z_j(\boldsymbol{x}) = \tanh(h_j(\boldsymbol{x})) \in (-1, 1)$, implying that each Chebyshev term $T_n(z_j)$ is bounded within $[-1, 1]$. Given $z_j$, the values $T_n(z_j)$ are deterministic, so each product $w_{j,1,n}^{(2)} \cdot T_n(z_j)$ is a Gaussian random variable with zero mean and variance $T_n(z_j)^2$. Therefore, conditioned on the values of $\{z_j\}$, the output $f$ is normally distributed,
\begin{equation}
f \mid \{z_j\} \sim \mathcal{N}\left(0,\; \sum_{j=1}^{N} \sum_{n=0}^{k} T_n(z_j)^2\right).
\end{equation}

Since the variables $z_j$ themselves are random, the marginal distribution of $f$ is a Gaussian mixture, meaning that $f$ has a distribution whose variance depends on the particular realization of $\{z_j\}$. In expectation, the variance of $f$ is given by,
\begin{equation}
\mathbb{V}[f] = \mathbb{E}_{\{z_j\}}\left[ \sum_{j=1}^{N} \sum_{n=0}^{k} T_n(z_j)^2 \right],
\end{equation}
that means the output $f$ has zero mean, i.e., $\mathbb{E}[f] = 0$, and a variance that depends on the input coordinate values $\tilde{x}_i = \tanh(x_i)$, the network width $N$, the Chebyshev polynomial degree $k$, and how the nonlinearity $\tanh$ transforms the Gaussian pre-activations $h_j$.
In practice, due to the summation over many (weakly dependent) terms, the output $f$ will approximately follow a Gaussian distribution by the Central Limit Theorem \cite{mood1950introduction, ross2009probability, bishop2006pattern}, especially as the number of neurons $N$ increases. For notational simplicity in the following derivations, we use $\boldsymbol{x}$ and $\boldsymbol{x}'$ to denote the full input vectors, corresponding to $\boldsymbol{x}_i$ and $\boldsymbol{x}_j$ in Eq.~\eqref{Eq:K_NTK_ij}, while $x_i$ or $x_j$ refer to individual coordinates within these vectors.

\begin{theorem}\label{Theorem:NTKcKAN}
Let $f(\boldsymbol{x}; \boldsymbol{\theta})$ denote the output of a cKAN with one hidden layer of width $N$, where all coefficients are initialized as independent standard normal random variables. The expected NTK between two inputs $\boldsymbol{x}$ and $\boldsymbol{x}'$ is given by,
\begin{equation}
\mathbb{E}[\boldsymbol{K}_{\text{ntk}}(\boldsymbol{x}, \boldsymbol{x}')] = N \cdot \left[
\sum_{n=0}^k C_n(\boldsymbol{x}, \boldsymbol{x}') + \sum_{i=1}^d \sum_{n=0}^k T_n(\tilde{x}_i) \cdot T_n(\tilde{x}'_i) \cdot D(\boldsymbol{x},\boldsymbol{x}')
\right],
\end{equation}
where $T_n(\cdot)$ denotes the Chebyshev polynomial of degree $n$, and $\tilde{x}_i = \tanh(x_i)$ represents the transformed input coordinate. The term $C_n(\boldsymbol{x}, \boldsymbol{x}')$ captures the correlation between the Chebyshev features of hidden activations and is defined as,
\begin{equation}
C_n(\boldsymbol{x}, \boldsymbol{x}') = \mathbb{E}[T_n(\tanh(h)) \cdot T_n(\tanh(h'))],
\end{equation}
where the pair $(h, h')$ follows a bivariate normal distribution with zero mean and covariance matrix,
\begin{equation}
\begin{bmatrix}
    \sigma^2(\boldsymbol{x}) & \rho(\boldsymbol{x}, \boldsymbol{x}') \\
    \rho(\boldsymbol{x}, \boldsymbol{x}') & \sigma^2(\boldsymbol{x}')
\end{bmatrix},
\end{equation}
and the second term, $D(\boldsymbol{x}, \boldsymbol{x}')$, accounts for the contribution of gradients through the activation function and is given by,
\begin{equation}
D(\boldsymbol{x}, \boldsymbol{x}') = \mathbb{E} \left[
    (1 - \tanh^2(h)) (1 - \tanh^2(h')) \sum_{m=0}^k T_m'(\tanh(h)) \cdot T_m'(\tanh(h'))
\right],
\end{equation}
where $(h, h')$ has the same distribution as above.
\end{theorem}\noindent
The proof of Theorem \ref{Theorem:NTKcKAN} can be found in \ref{App.Proof1}.
\begin{remark}
The expected NTK increases linearly with the number of hidden neurons $N$. It is composed of two main components. The first part, known as the second-layer kernel, involves the terms $C_n$, which measure how similar two inputs are after passing through the nonlinear activation function $\tanh$ and being expanded using Chebyshev polynomials. The second part, referred to as the first-layer kernel, involves the term $D$, which captures how sensitive the network output is to changes in the first-layer coefficients; this sensitivity is expressed through derivatives of the Chebyshev polynomials. Importantly, both components depend only on the input data and their statistical properties, not on the specific random initialization of the network coefficients.
\end{remark}

As training progresses, the parameter vector $\boldsymbol{\theta}(\tau)$ deviates from its initial random value due to updates from gradient descent. Understanding how much the NTK changes over time, referred to as NTK drift, is crucial for analyzing the learning dynamics of finite-width cKANs.

\begin{theorem}\label{Theorem:NTKConv}
Let $f(\boldsymbol{x}; \boldsymbol{\theta})$ be a cKAN parameterized by weights $\boldsymbol{\theta} \in \mathbb{R}^P$, and let the empirical NTK matrix at time $\tau$ be defined as,
\begin{equation}
    \boldsymbol{K}_{\text{ntk}}^{(\tau)}(\boldsymbol{x}, \boldsymbol{x}') := \left\langle \nabla_{\boldsymbol{\theta}} f(\boldsymbol{x}; \boldsymbol{\theta}(\tau)), \nabla_{\boldsymbol{\theta}} f(\boldsymbol{x}'; \boldsymbol{\theta}(\tau)) \right\rangle.
\end{equation}
Assume the following:
\begin{itemize}
    \item[(I)] 
    The parameters evolve under gradient flow,
    \begin{equation}
        \frac{d \boldsymbol{\theta}}{d\tau} = -\nabla_{\boldsymbol{\theta}} \mathcal{L}(\boldsymbol{\theta}(\tau)),
    \end{equation}
    where $\mathcal{L}$ is a smooth loss function defined in Eq.~\eqref{Eq.SmoothLoss}.
    \item[(II)]
    The gradients and Hessians of the network outputs are uniformly bounded,
    \begin{equation}
        \sup_{\boldsymbol{\theta}, \boldsymbol{x}} \left\| \nabla_{\boldsymbol{\theta}} f(\boldsymbol{x}; \boldsymbol{\theta}) \right\| \leq B_1, \quad \sup_{\boldsymbol{\theta}, \boldsymbol{x}} \left\| \nabla_{\boldsymbol{\theta}}^2 f(\boldsymbol{x}; \boldsymbol{\theta}) \right\| \leq B_2,
    \end{equation}
    for some constants $B_1, B_2 > 0$.
\end{itemize}
Then, the NTK matrix remains close to its initialization for a finite training time $\tau$,
\begin{equation}
    \left\| \boldsymbol{K}_{\text{ntk}}(\tau) - \boldsymbol{K}_{\text{ntk}}(0) \right\| \leq C \cdot \left\| \boldsymbol{\theta}(\tau) - \boldsymbol{\theta}(0) \right\|,
\end{equation}
where $C = 2 B_1 B_2$. In particular, if the parameter change $\left\| \boldsymbol{\theta}(\tau) - \boldsymbol{\theta}(0) \right\|$ vanishes as the width $N \to \infty$, the NTK remains asymptotically constant during training.
\end{theorem}\noindent
The proof of Theorem \ref{Theorem:NTKConv} can be found in \ref{App.Proof2}.

\begin{remark}
Theorem~\ref{Theorem:NTKConv} implies that under gradient flow and boundedness conditions, the NTK matrix of a cKAN model remains nearly constant throughout training, especially in the infinite-width limit. This stability is a key assumption in many theoretical analyses of neural network training dynamics and justifies using the NTK at initialization to approximate the learning behavior of the model. In particular, cKAN, which is based on Chebyshev polynomial expansions, satisfies these conditions due to its smooth structure and bounded derivatives.
\end{remark}

\subsection{Step II: NTK for cPIKANs}


This section aims to derive the NTK formulation for the cPIKAN framework. Consider a well-posed partial differential equation defined over a bounded domain $\Omega \subset \mathbb{R}^d$, given by,
\begin{equation}
\mathcal{N}[u(\boldsymbol{x})](\boldsymbol{x}) = h(\boldsymbol{x}), \quad \boldsymbol{x} \in \Omega
\end{equation}
\begin{equation}
u(\boldsymbol{x}) = g(\boldsymbol{x}), \quad \boldsymbol{x}\in \partial\Omega
\end{equation}
where $\mathcal{N}$ is a differential operator, and $u: \Omega \rightarrow \mathbb{R}$ is the unknown function to be learned, with $\boldsymbol{x} = (x_1, x_2, \dots, x_d)$. For time-dependent problems, time $t$ is included as an additional coordinate in $\boldsymbol{x}$, and $\Omega$ represents the combined space-time domain. In such cases, the initial condition is treated as a special case of a Dirichlet boundary condition.
Following the vanilla physics-informed formulation, we approximate the solution $u$ using a deep neural network $u(\boldsymbol{x}; \boldsymbol{\theta})$, where $\boldsymbol{\theta}$ denotes the trainable parameters of the model. The PDE residual at any input $\boldsymbol{x}$ is defined as,
\begin{equation}
r(\boldsymbol{x}; \boldsymbol{\theta}) := \mathcal{N}[u](\boldsymbol{x}; \boldsymbol{\theta}) - h(\boldsymbol{x}).
\end{equation}

The network parameters are optimized by minimizing a loss function composed of two parts: the data loss and the residual loss. Specifically, the total loss is given by,
\begin{equation}\label{Eq.Loss_PINN}
\mathcal{L}(\boldsymbol{\theta}) = \underbrace{\frac{1}{2} \sum_{i=1}^{N_d} \left| u(\boldsymbol{x}_i^d; \boldsymbol{\theta}) - g(\boldsymbol{x}_i^d) \right|^2}_{\mathcal{L}_d(\boldsymbol{\theta})} + \underbrace{\frac{1}{2} \sum_{i=1}^{N_r} \left| r(\boldsymbol{x}_i^r; \boldsymbol{\theta}) \right|^2}_{\mathcal{L}_r(\boldsymbol{\theta})}
\end{equation}
where \(\mathcal{L}_d(\boldsymbol{\theta})\) is the loss from data or boundary/initial conditions, and \(\mathcal{L}_r(\boldsymbol{\theta})\) is the residual loss from the PDE.
Here, $N_d$ and $N_r$ denote the number of training points sampled from the boundary and/or initial conditions and the interior of the domain, respectively, with corresponding datasets defined as,
\begin{equation}\label{Eq.TrainingData}
\mathcal{D}_d = \left\{ (\boldsymbol{x}_i^d, g(\boldsymbol{x}_i^d)) \right\}_{i=1}^{N_d}, \quad \mathcal{D}_r = \left\{ (\boldsymbol{x}_i^r, h(\boldsymbol{x}_i^r)) \right\}_{i=1}^{N_r},
\end{equation}
where $\boldsymbol{x}_i^d \in \partial\Omega$ and $\boldsymbol{x}_i^r \in \Omega$ indicate the sampled input locations for the data and residual loss terms, respectively.

To optimize the neural network, we seek the parameter vector $\boldsymbol{\theta}$ that minimizes the loss function $\mathcal{L}(\boldsymbol{\theta})$. A widely used approach for this task is gradient descent, which iteratively updates the parameters in the direction of the negative gradient of the loss.
When the learning rate is taken to be infinitesimally small, the discrete update rule of gradient descent can be approximated by a continuous-time process known as gradient flow \cite{bottou2018optimization, jacot2018neural}. In this regime, the evolution of the parameters is governed by the differential equation,
\begin{equation}\label{Eq.Gloss}
\frac{d\boldsymbol{\theta}(\tau)}{d\tau} = -\nabla_{\boldsymbol{\theta}} \mathcal{L}(\boldsymbol{\theta}(\tau)) = -\nabla_{\boldsymbol{\theta}} \:\left(\mathcal{L}_d(\boldsymbol{\theta}(\tau)) + \mathcal{L}_r(\boldsymbol{\theta}(\tau))\right),
\end{equation}
where $\tau$ is training time. This equation describes how the parameters $\boldsymbol{\theta}$ evolve smoothly along the direction of steepest descent of the loss function.
\begin{lemma}\label{Lemma:NTK}
Let the training data be given by Eq.~\eqref{Eq.TrainingData}.
Under the gradient flow dynamics defined in Eq.~\eqref{Eq.Gloss}, the evolution of the network outputs $u(\boldsymbol{x}^d_i; \boldsymbol{\theta}(\tau))$ and $\mathcal{N}[u](\boldsymbol{x}^r_i; \boldsymbol{\theta}(\tau))$ with respect to the training time $\tau$ follows,
\begin{equation}
\frac{d}{d\tau} 
\begin{bmatrix} 
u(\boldsymbol{x}^d_i; \boldsymbol{\theta}(\tau)) \\ 
\mathcal{N}[u](\boldsymbol{x}^r_i; \boldsymbol{\theta}(\tau)) 
\end{bmatrix}
=
- 
\begin{bmatrix}
K_{uu}(\tau) & K_{ur}(\tau) \\
K_{ru}(\tau) & K_{rr}(\tau)
\end{bmatrix}
\begin{bmatrix}
u(\boldsymbol{x}^d_i; \boldsymbol{\theta}(\tau)) - g(\boldsymbol{x}^d_i) \\
\mathcal{N}[u](\boldsymbol{x}^r_i; \boldsymbol{\theta}(\tau)) - h(\boldsymbol{x}^r_i)
\end{bmatrix},
\end{equation}
where the kernel blocks $K_{uu}(\tau)$, $K_{ur}(\tau)$, $K_{ru}(\tau)$, and $K_{rr}(\tau)$ are components of the empirical NTK, given by inner products of the parameter gradients of the network outputs and residuals.
Specifically,
 \begin{equation}
    \begin{array}{l}
       (K_{uu}(\tau))_{i,j} =  \displaystyle{\left\langle \frac{\partial u(\boldsymbol{x}^d_i; \boldsymbol{\theta}(\tau))}{\partial \boldsymbol{\theta}}, \frac{\partial u(\boldsymbol{x}^d_j; \boldsymbol{\theta}(\tau))}{\partial \boldsymbol{\theta}} \right\rangle },  \\[4mm]
        (K_{rr}(\tau))_{i,j} = \displaystyle{\left\langle \frac{\partial \mathcal{N}[u](\boldsymbol{x}^r_i; \boldsymbol{\theta}(\tau))}{\partial \boldsymbol{\theta}}, \frac{\partial \mathcal{N}[u](\boldsymbol{x}^r_j; \boldsymbol{\theta}(\tau))}{\partial \boldsymbol{\theta}} \right\rangle },\\[4mm]
        (K_{ru}(\tau))_{i,j} =\displaystyle{\left\langle \frac{\partial \mathcal{N}[u](\boldsymbol{x}^r_i; \boldsymbol{\theta}(\tau))}{\partial \boldsymbol{\theta}}, \frac{\partial u(\boldsymbol{x}^d_j; \boldsymbol{\theta}(\tau))}{\partial \boldsymbol{\theta}} \right\rangle },
    \end{array}
\end{equation}
where $\langle \cdot \rangle$ denotes the inner product taken over all neural network parameters $\boldsymbol{\theta} = \{c_n\}_{n=0}^{k}$, which represent the coefficients in the Chebyshev polynomial expansion.
\end{lemma}
The proof of Lemma \ref{Lemma:NTK} follows the same reasoning as the NTK analysis for PINNs presented in \cite{wang2022and}.
Accordingly, the matrix,
\begin{equation}\label{Eq:NTK1}
    \boldsymbol{K}_{\text{ntk}} (\tau) = 
    \begin{bmatrix}
K_{uu}(\tau) & K_{ur}(\tau) \\
K_{ru}(\tau) & K_{rr}(\tau)
\end{bmatrix},
\end{equation}
is referred to as the NTK of the cPIKAN. 
It characterizes the sensitivities of both the cKAN outputs $u(\boldsymbol{x}^d_i; \boldsymbol{\theta})$ and the physics-informed residuals $\mathcal{N}[u](\boldsymbol{x}^r_i; \boldsymbol{\theta})$ with respect to the model parameters $\boldsymbol{\theta}$.
These kernels capture how perturbations in the network parameters influence the model outputs and their corresponding residuals across the training data.
The evolution equation,
\begin{equation}\label{Eq:NTK100}
\frac{d \boldsymbol{\psi}(\tau)}{d\tau} = - \boldsymbol{K}_{\text{ntk}}(\tau) \cdot (\boldsymbol{\psi}(\tau) - \boldsymbol{\mathcal{G}})
\end{equation}
describes a  linearized training dynamic  that holds broadly for physics-informed models under gradient flow, regardless of the specific neural network architecture used (e.g., fully connected networks, cKANs, Fourier networks).
Here, $\boldsymbol{\psi}(\tau)$ is the stacked vector of network outputs corresponding to different components of the problem, including PDE residuals, initial and boundary conditions, and possibly data observations; $\boldsymbol{\mathcal{G}}$ contains the corresponding target values; and $\boldsymbol{K}_{\text{ntk}}(\tau)$ is the NTK matrix, given by,
\begin{equation}
\boldsymbol{K}_{\text{ntk}}(\tau) =
\begin{bmatrix}
    K_{\psi_1, \psi_1}(\tau) & K_{\psi_1, \psi_2}(\tau) & \cdots & K_{\psi_1, \psi_m}(\tau) \\
    K_{\psi_2, \psi_1}(\tau) & K_{\psi_2, \psi_2}(\tau) & \cdots & K_{\psi_2, \psi_m}(\tau) \\
    \vdots & \vdots & \ddots & \vdots \\
    K_{\psi_m, \psi_1}(\tau) & K_{\psi_m, \psi_2}(\tau) & \cdots & K_{\psi_m, \psi_m}(\tau)
\end{bmatrix}, 
\end{equation}
where each block $K_{\psi_i, \psi_j}(\tau)$ encodes the inner product of gradients of the network outputs $\psi_i$ and $\psi_j$ with respect to the model parameters.
This formulation is flexible and  extends naturally to a wide class of PDEs, including time-dependent and nonlinear problems, with varied boundary and initial conditions. It provides a unified view of training dynamics across different network types and physics constraints.



\section{Results \& Discussion}

In this section, we present the results of four numerical experiments designed to evaluate the accuracy, computational efficiency, and learning dynamics of cPIKAN architectures, with particular emphasis on the eigenvalue behavior of NTK matrices under various settings. Since the computational domains in our experiments extend beyond the unit scale, we employ the scaled cPIKAN method as proposed in \cite{mostajeran2025scaled}. Each subsection targets a distinct aspect of model behavior or optimization strategy. Specifically, the experiments include: (I) a comparison between PINN and cPIKAN on the diffusion equation, (II) an evaluation of different optimization methods and model variants on the Helmholtz equation, (III) an analysis of NTK structure in temporally decomposed subdomains for the Allen–Cahn equation, and (IV) the evolution of learned hyperparameters in the context of a forced vibration problem. For each case, we report the network architecture, number of trainable parameters $\vert \boldsymbol{\theta}\vert$, relative $\mathcal{L}^2$ error (defined as $(\Vert u_{\text{pred}} - u_{\text{exact}}\Vert_2)/(\Vert u_{\text{exact}}\Vert_2)$), and the average computation time per training iteration, measured on a workstation equipped with an NVIDIA RTX 6000 Ada GPU. 
In addition, we provide visualizations of the ground truth, predicted solution, and absolute error over the computational domain, along with training curves for the loss and relative $\mathcal{L}^2$ error per iteration. 
The eigenvalue spectra of the NTK matrices, denoted by $\lambda(K_{\psi_i, \psi_j}(\tau))$, are also presented to offer further insights into convergence and learning dynamics.
We note that in Experiments~\ref{Exam.DiffEqu}–\ref{Exam.AC}, the NTK matrix is defined according to Lemma~\ref{Lemma:NTK} and Eq.~\eqref{Eq:NTK1}. However, in Experiment~\ref{Exam:Vib}, due to the presence of distinct boundary and initial conditions, the formulation of the NTK differs and is discussed separately within that example.

\subsection{Experiment I: PINN and cPIKAN (Diffusion Equation)}\label{Exam.DiffEqu}


In this example, we aim to investigate and compare the behavior of the eigenvalues of the NTK matrix for two approaches: the scaled versions of the PINN and the cPIKAN, as introduced in \cite{mostajeran2025scaled}.
To this end, we consider the one-dimensional diffusion equation defined as,
\begin{equation}
    \begin{array}{ll}
        u_{t}(x,t) - D\: u_{xx}(x,t) = 0, &  x \in [-6, 6], t\in (0, 1],\\[3mm]
        u(-6,t) = u(6,t) = 0, &t\in (0, 1] ,\\[3mm]
        u(x, 0) = \sin(\pi x), &x \in [-6, 6],
\end{array}
\end{equation}
 where $D$ is the diffusion coefficient. 
 The ground truth solution for this problem is known and given $u(x, t) = \sin(\pi x) \exp(-\pi^2 D t)$. 
 This setup allows us to explore how the NTK spectrum evolves during training and how it reflects the learning dynamics and inductive biases of the two methods.

\begin{table}[!h]
\centering
\caption{\label{Tab.HeatSetting} 
Comparison of network configurations, number of trainable parameters $\vert \boldsymbol{\theta}\vert$, relative $\mathcal{L}^2$ errors (RE), and average computation time per iteration (in milliseconds) for solving the diffusion equation in Experiment~\ref{Exam.DiffEqu} with diffusion coefficient $D = 0.1$. All models use equal residual and data loss weights. 
}
\renewcommand{\arraystretch}{1.2}
\setlength{\tabcolsep}{12pt}
\begin{tabular}{c|ccc|cc}
\toprule
Method & $(N_l, N_n, k)$ & $\vert \boldsymbol{\theta}\vert$& \((N_{\text{r}}, N_{\text{d}})\) & RE & Time \\
\midrule
cPIKAN & (2, 8, 5) & 462&(2000, 800) &$\mathbf{2.18 \times 10^{-3}}$ & 6 \\
\midrule
\midrule
\multicolumn{6}{l}{Parameter-based analysis} \\
\midrule
PINN-I & (2, 19, -) & 456 &(2000, 800) & $8.62 \times 10^{-3}$& 2\\
\midrule
\midrule
\multicolumn{6}{l}{ Computation-time-based comparison} \\
\midrule
PINN-II & (5, 100, -) & 40802 & (40000, 800) &$1.46 \times 10^{-2}$& 10\\
\bottomrule
\end{tabular}
\end{table}

Before analyzing the eigenvalues of the NTK matrices, we present a summary of the training configurations and performance metrics for the compared models in Table~\ref{Tab.HeatSetting}. The PINN architecture is evaluated in two settings: PINN-I is configured to match the number of trainable parameters with cPIKAN (parameter-based comparison), while PINN-II is designed to have a similar computation time per iteration (time-based comparison). As shown, cPIKAN achieves the lowest relative $\mathcal{L}^2$ error ($2.18 \times 10^{-3}$) while maintaining a moderate computational cost (6 ms per iteration), offering approximately 4
$\times $ higher accuracy than PINN-I and nearly 7$\times $ higher accuracy than PINN-II, despite significantly fewer parameters and lower sampling requirements. These results highlight the superior efficiency and precision of cPIKAN in solving the diffusion problem.

\begin{figure}[!h]
    \centering
    \includegraphics[width=0.99\linewidth]{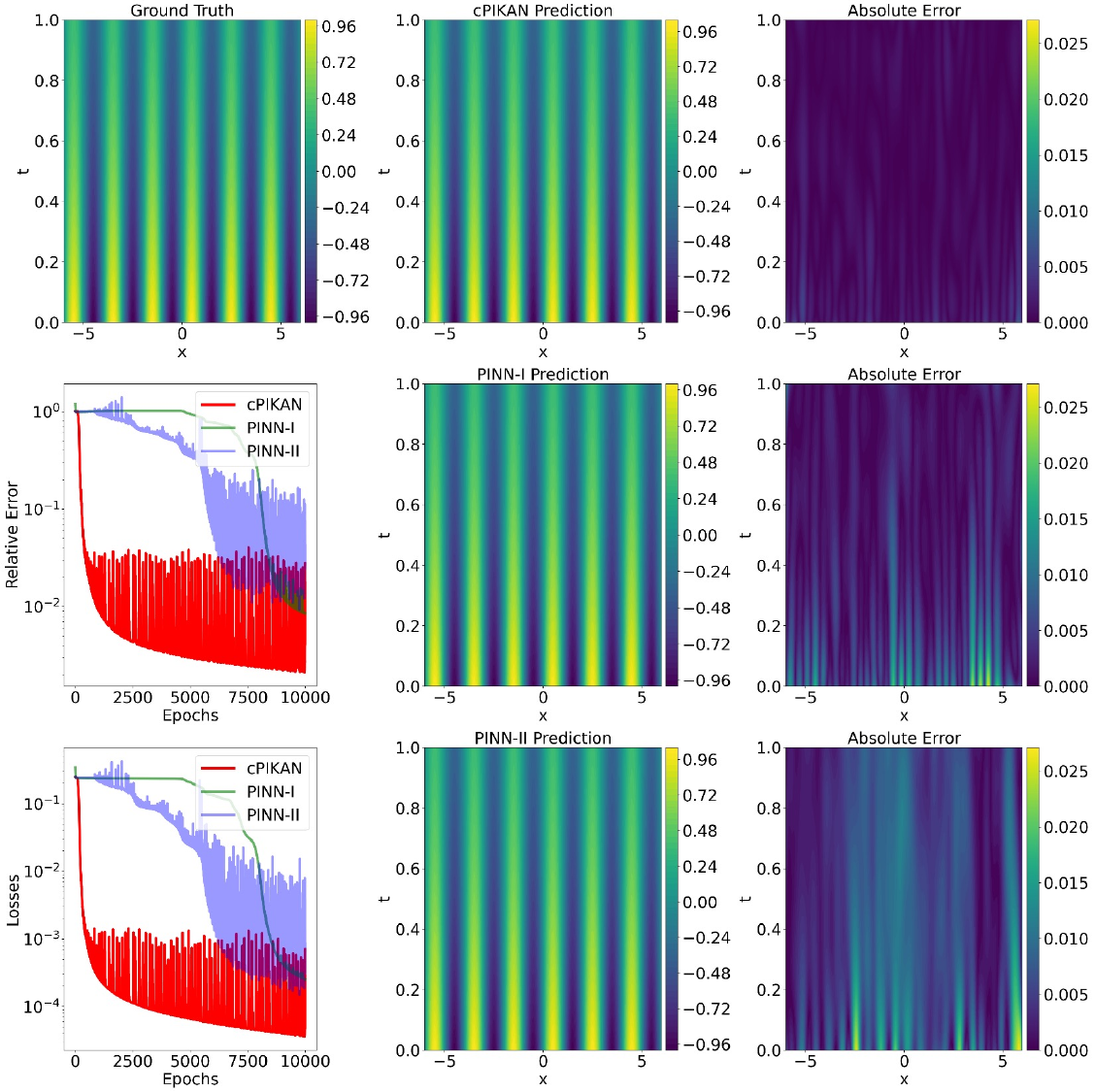}
    \caption{Comparison of the predicted solutions for the diffusion equation in Experiment~\ref{Exam.DiffEqu}, demonstrating that  cPIKAN  yields significantly higher accuracy, with a maximum absolute error of only $7.52 \times 10^{-3}$, compared to PINN-I and PINN-II, whose errors exceed $2.5 \times 10^{-2}$. cPIKAN also exhibits faster convergence, lower final relative $\mathcal{L}^2$ error, and more stable training dynamics. These findings highlight the superior performance of cPIKAN in both predictive accuracy and optimization efficiency for solving the diffusion problem.}
    \label{fig:NTK_Diff_RelE}
\end{figure}

In Fig.~\ref{fig:NTK_Diff_RelE}, we evaluate and compare the performance of cPIKAN, PINN-I, and PINN-II in solving the one-dimensional diffusion equation. The predicted solutions and corresponding absolute error distributions indicate that cPIKAN achieves the highest accuracy, with a maximum absolute error of $7.52 \times 10^{-3}$, which is substantially lower than that of PINN-I ($2.53 \times 10^{-2}$) and PINN-II ($2.70 \times 10^{-2}$). The plots in the bottom-left of the figure depict the evolution of the relative $\mathcal{L}^2$ error and the total training loss as functions of training epochs. These curves show that cPIKAN not only attains the smallest final relative $\mathcal{L}^2$ error but also demonstrates faster convergence and greater training stability compared to the standard PINN configurations. This highlights the effectiveness of cPIKAN in both accuracy and optimization dynamics in solving Experiment~\ref{Exam.DiffEqu}.

\begin{figure}[!h]
    \centering
    \includegraphics[width=0.99\linewidth]{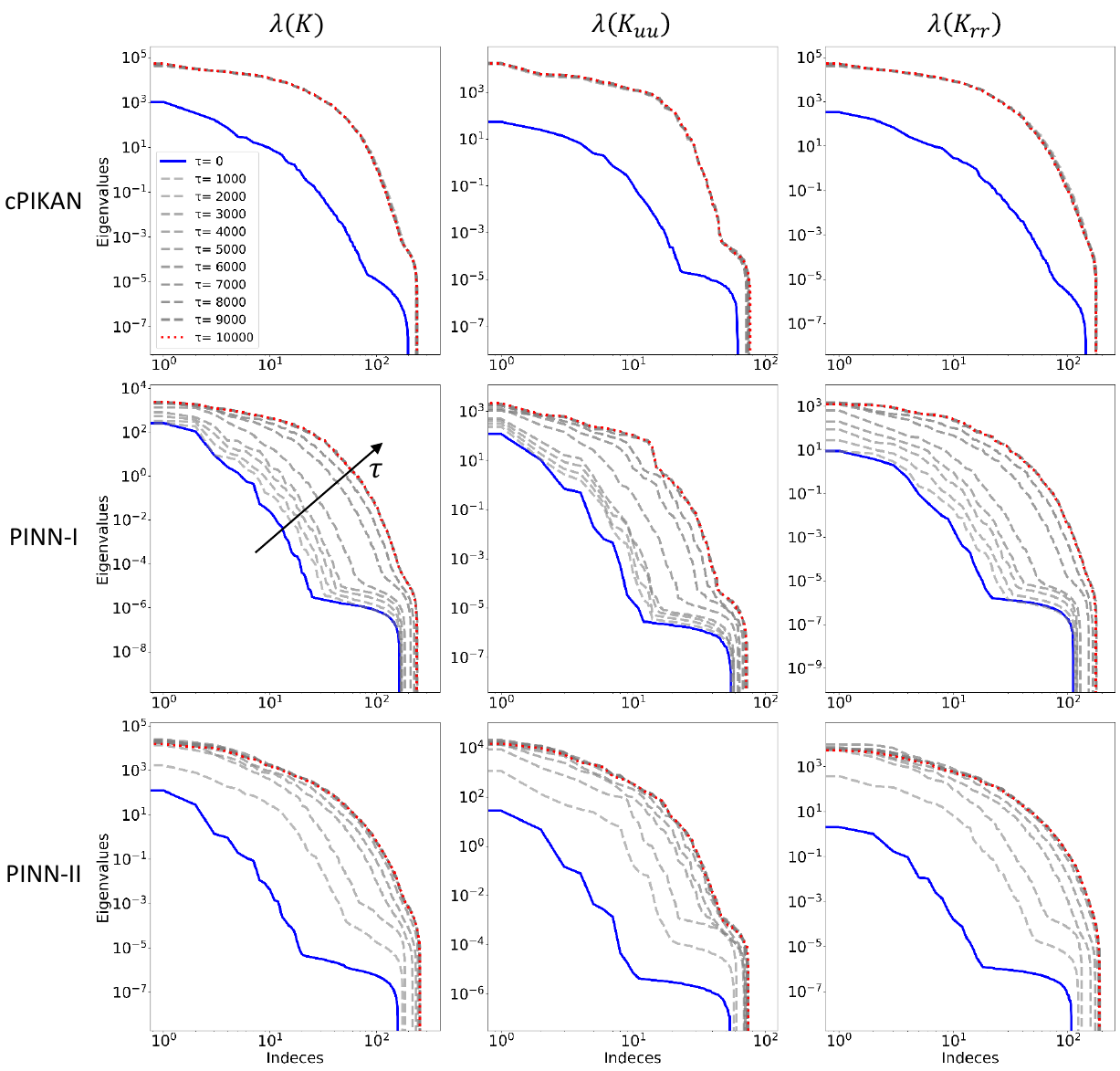}
    \caption{Evolution of the NTK eigenvalue spectra during training for the diffusion equation in Experiment~\ref{Exam.DiffEqu}. The NTK spectrum in cPIKAN gradually converges during training, indicating stable learning dynamics and effective optimization. In contrast, the spectra for PINN-I and PINN-II remain dispersed, suggesting unstable behavior and poor information flow. These differences highlight the improved convergence and robustness of the cPIKAN model.
    }
    \label{fig:NTK_Diff}
\end{figure}

Figure \ref{fig:NTK_Diff} illustrates the temporal evolution of the NTK eigenvalue spectra during training for three architectures: cPIKAN, PINN-I, and PINN-II. Each columns represent the eigenvalues of the full NTK matrix $K_{\text{ntk}}$, the data-data submatrix $K_{\text{uu}}$, and the residual-residual submatrix $K_{\text{rr}}$, respectively. For all models, the spectra are plotted over a range of training iterations from $\tau = 0$ (blue) to $\tau = 10^4$ (red).
The plots reveal distinct spectral dynamics across models. Notably, cPIKAN maintains a significantly broader and more persistent spectrum throughout training. This indicates enhanced expressivity and better signal propagation through both data and residual losses, facilitating stable convergence. In contrast, the spectra of PINN-I and PINN-II rapidly collapse towards lower magnitudes, especially in the residual blocks. This decay reflects limited capacity to represent gradient information and suggests poor learning signal transmission in deeper iterations.
Moreover, the eigenvalue spectrum of $K_{\text{rr}}$ for cPIKAN retains higher magnitudes across more modes compared to PINN-I and PINN-II. This implies that cPIKAN better preserves the diversity of learning directions in the residual space, which contributes to its superior performance in both accuracy and convergence behavior observed in earlier figures.

\clearpage

\subsection{Experiment II: Different Optimizations (Helmholtz Equation)}\label{Exam.HelmHEqu}


In this example, we assess the performance of various models, including the scaled version of cPIKAN, standard PINN, and the B-spline-based PINN (bPIKAN) introduced in \cite{mostajeran2025scaled}, for solving the two-dimensional Helmholtz equation with homogeneous Dirichlet boundary conditions,
\begin{equation}
\begin{array}{ll}
u_{xx}(x,y) + u_{yy}(x,y) + \kappa^2 u(x,y) = f(x, y), & (x, y) \in \Omega,\\[3mm]
u(x,y) = 0, &(x, y) \in \partial \Omega ,
\end{array} 
\end{equation}
where the domain is $\Omega = [-4, 4]^2$. The source term is defined as 
$
f(x, y) = \left(\kappa^2 - (a_1^2 + a_2^2)\pi^2\right) \sin(a_1 \pi x) \sin(a_2 \pi y),$
with the corresponding exact solution 
$
u(x, y) = \sin(a_1 \pi x) \sin(a_2 \pi y),
$
which satisfies the prescribed boundary conditions. The parameters $a_1$, $a_2$, and $\kappa$ control the oscillatory behavior of the solution. In this test, we use $(a_1, a_2, \kappa) = (1.0, 1.0, 1.0)$, yielding a smooth yet non-trivial benchmark. In addition to comparing model architectures, we also investigate the effect of different optimization strategies (e.g., ADAM, LBFGS, a hybrid scheme) on the NTK spectra and training dynamics of cPIKAN.
The LBFGS optimizer is configured with a strong Wolfe line search to promote stable and efficient convergence.

\begin{table}[!h]
\centering
\caption{\label{Tab.HelmHSetting} 
Comparison of network configurations, total number of trainable parameters $\vert \boldsymbol{\theta}\vert$, relative $\mathcal{L}^2$ errors (RE), and per-iteration computational time (in milliseconds) for solving the Helmholtz equation described in Experiment~\ref{Exam.HelmHEqu}. All models are trained with equal residual and data loss weights and use $N_{\text{r}} = N_{\text{d}} = 4000$ training points. The first three rows correspond to models trained with the ADAM optimizer. The bottom section reports the performance of cPIKAN under alternative optimization strategies, including LBFGS and a hybrid ADAM$+$LBFGS scheme.  }
\renewcommand{\arraystretch}{1.2}
\setlength{\tabcolsep}{12pt}
\begin{tabular}{c|cc|cc}
\toprule
Method & $(N_l, N_n, k)$ & $\vert \boldsymbol{\theta}\vert$& RE & Time \\
\midrule
cPIKAN & (4,25,3) & 7800&$\mathbf{6.61 \times 10^{-2}}$ & 18 \\
\midrule
PINN & (4, 50, -) & 7850  & $2.65 \times 10^{0}$& 4\\
\midrule
bPIKAN & (4, 16, 3) \& 5 grid & 7344  &$1.67 \times 10^{0}$& 30\\
\midrule
\midrule
\multicolumn{5}{l}{Other Optimization of cPIKAN} \\
\midrule
LBFGS   &(4,25,3) & 7800&$\mathbf{4.76 \times 10^{-3}}$ & 18 \\
\midrule
ADAM $+$ LBFGS & (4,25,3) & 7800 &$\mathbf{5.03 \times 10^{-3}}$ & 18 \\
\bottomrule
\end{tabular}
\end{table}
\begin{figure}[!h]
    \centering
    \includegraphics[width=0.99\linewidth]{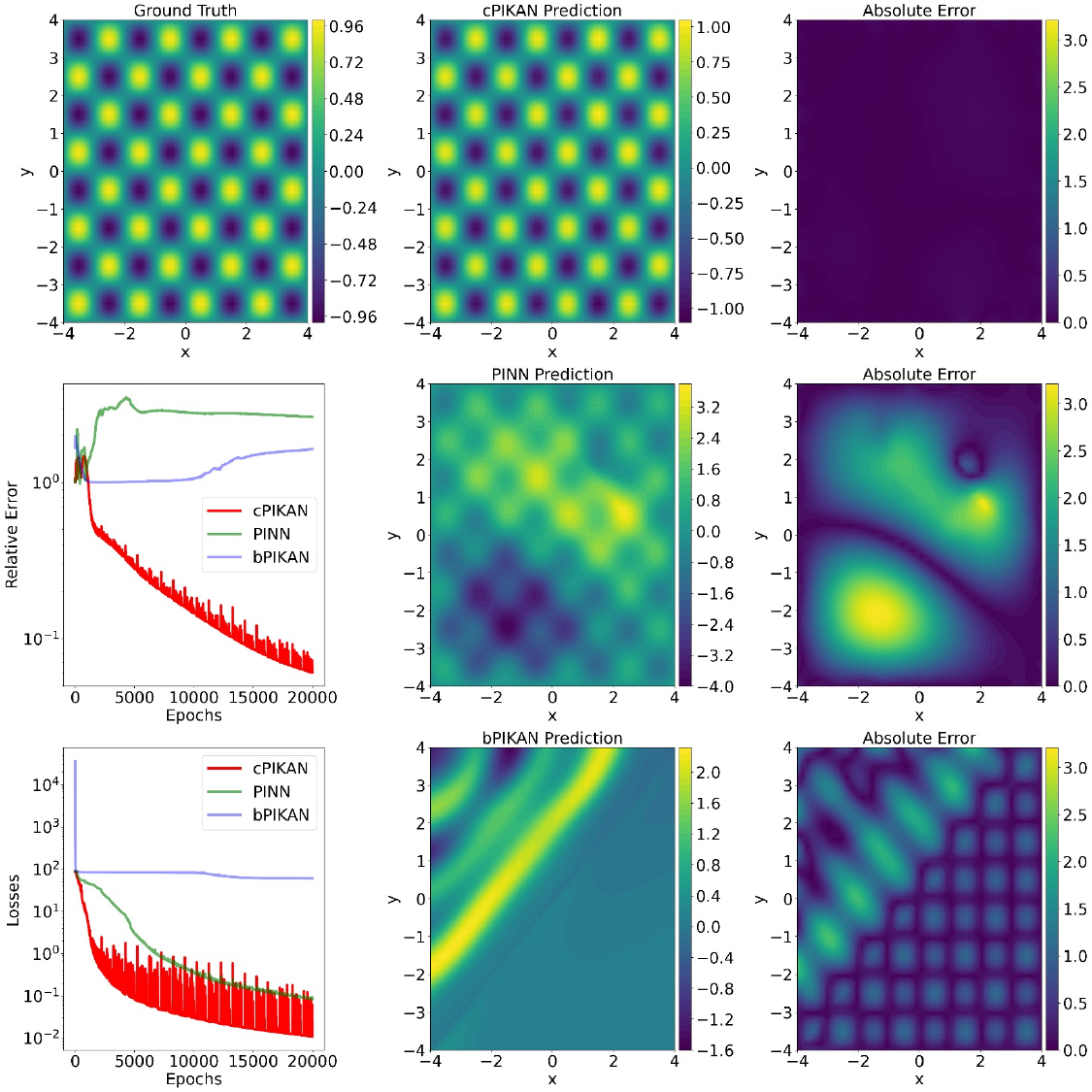}
    \caption{Comparison of the predicted solutions for the Helmholtz equation in Experiment~\ref{Exam.HelmHEqu}, showing that cPIKAN produces highly accurate predictions with minimal absolute error, closely matching the ground truth. PINN captures the overall structure but significantly misestimates the amplitude, while bPIKAN fails to approximate the solution correctly. The training curves further confirm these findings. cPIKAN achieves the lowest relative $\mathcal{L}^2$ error and loss with stable convergence, whereas PINN converges more slowly with higher error, and bPIKAN shows poor learning behavior and fails to converge effectively.
    }
    \label{fig:NTK_HelH_RelE}
\end{figure}
\begin{figure}[!h]
    \centering
    \includegraphics[width=0.99\linewidth]{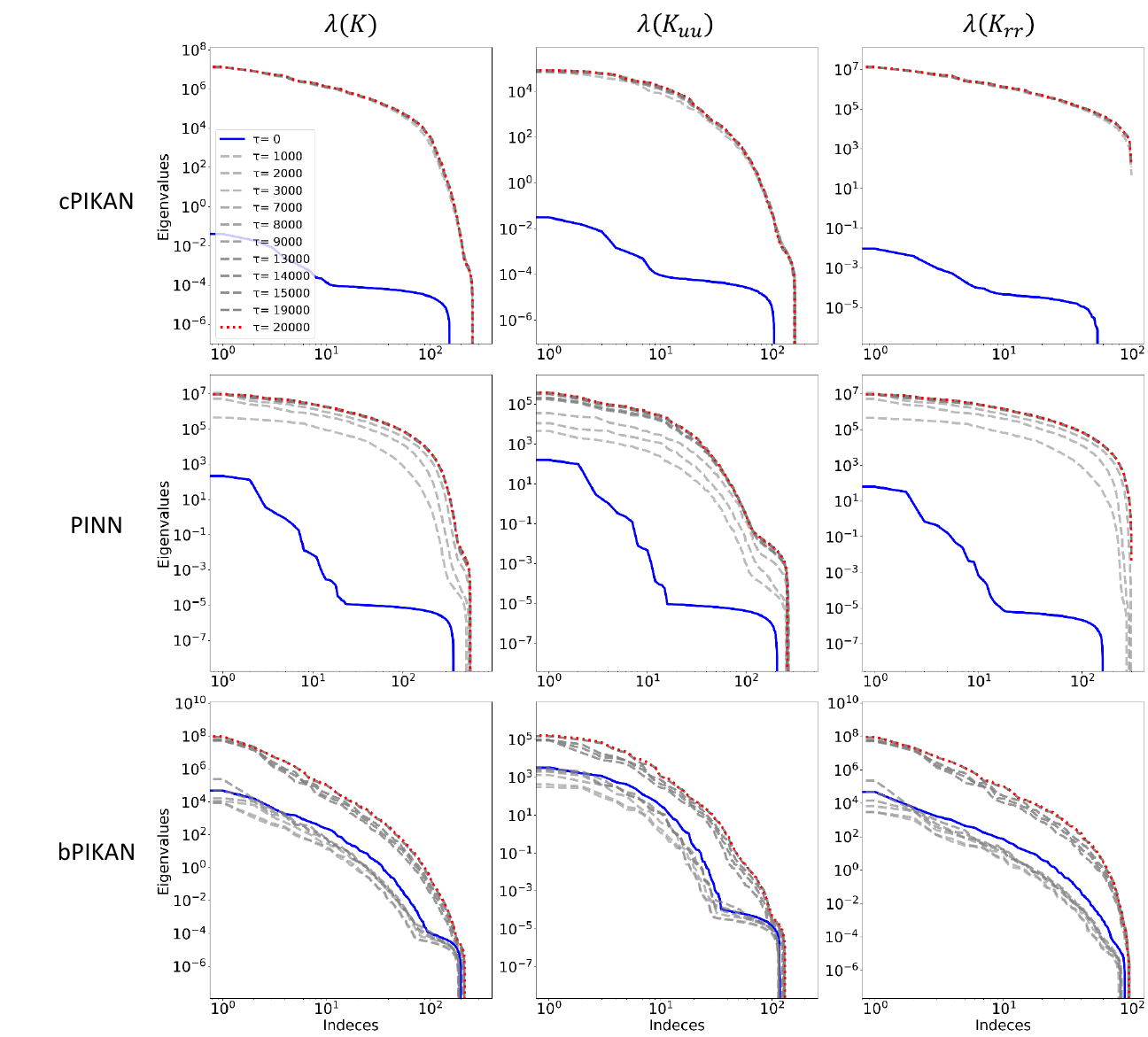}
    \caption{Evolution of the NTK eigenvalue spectra during training for the Helmholtz equation in Experiment~\ref{Exam.HelmHEqu}. The spectra for cPIKAN remain stable and gradually converge throughout training, reflecting well-conditioned dynamics and supporting its strong predictive performance. PINN shows delayed spectral stabilization, consistent with its slower convergence and moderate accuracy. In contrast, bPIKAN exhibits disordered and non-converging spectra, indicating unstable learning and poor approximation capability. 
    }
    \label{fig:NTK_HelH}
\end{figure}

\begin{figure}[!h]
    \centering
    \includegraphics[width=0.99\linewidth]{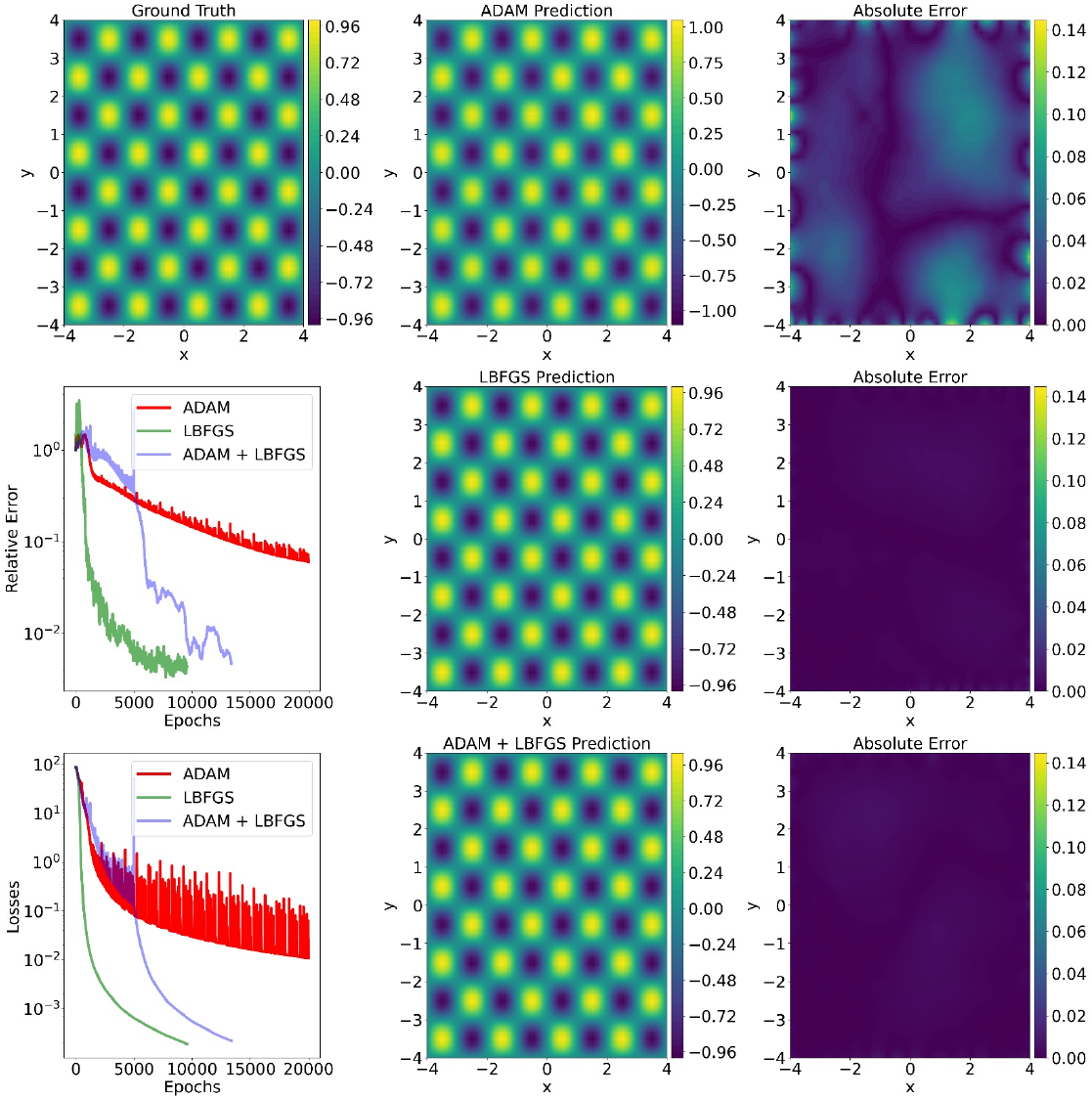}
    \caption{Comparison of predicted solutions for the Helmholtz equation using different optimization strategies in Experiment~\ref{Exam.HelmHEqu}. It is shown that combining ADAM with LBFGS leads to the most accurate and stable solution, achieving the lowest absolute and relative errors. LBFGS alone also performs well, while ADAM alone results in higher errors, especially near the domain center. Training curves confirm these observations. LBFGS converges fastest and most smoothly, followed by ADAM+LBFGS, whereas ADAM shows slower convergence and unstable loss behavior. These results highlight the importance of optimizer choice for achieving reliable and accurate training in cPIKAN.
    }
    \label{fig:NTK_HelH_OPT_RelE}
\end{figure}
\begin{figure}[!h]
    \centering
    \includegraphics[width=0.99\linewidth]{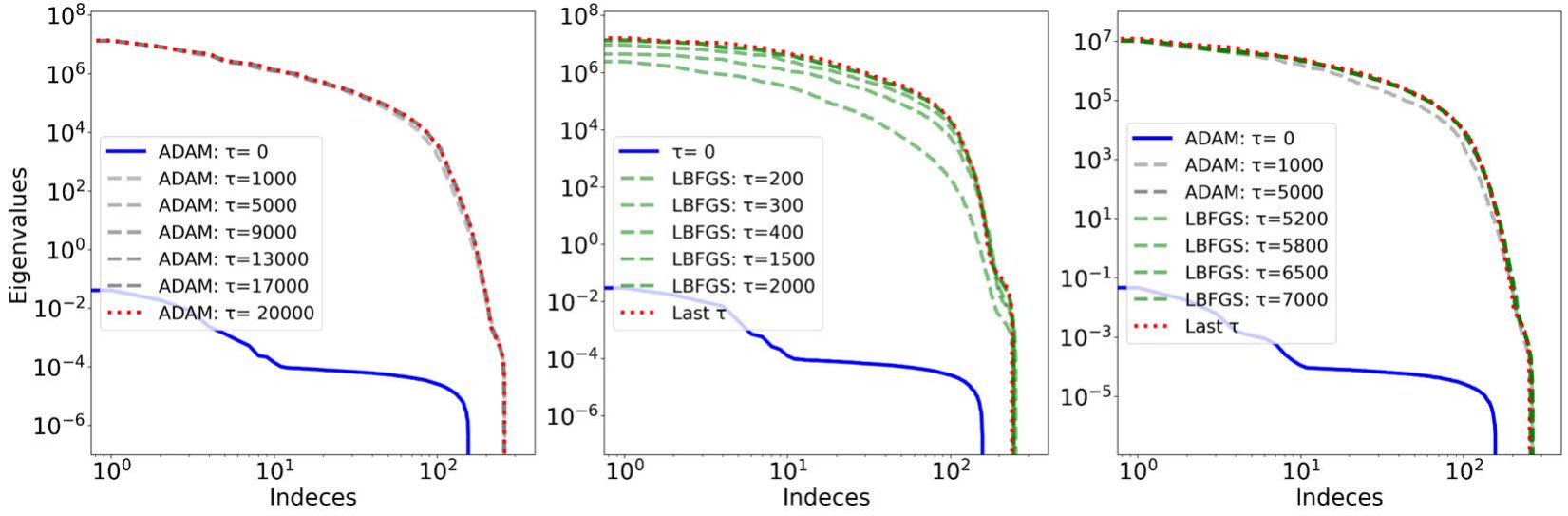}
    \caption{ Evolution of the NTK eigenvalue spectra during training for the Helmholtz equation in Experiment~\ref{Exam.HelmHEqu}, using three different optimization strategies.}
    \label{fig:NTK_HelH_OPT}
\end{figure}

Table~\ref{Tab.HelmHSetting} summarizes the performance of different models in solving the Helmholtz equation introduced in Experiment~\ref{Exam.HelmHEqu}, comparing their accuracy, computational efficiency, and network complexity.
 The first part of this table includes results for cPIKAN, PINN, and bPIKAN, all trained using the ADAM optimizer. Among these, cPIKAN achieves the best accuracy (relative $\mathcal{L}^2$ error of $6.61 \times 10^{-2}$) at a moderate computational cost of 18 ms per iteration. In contrast, standard PINN is significantly faster (4 ms per iteration) but exhibits much lower accuracy. The bPIKAN model yields a relatively high error, which may stem from its sensitivity to domain scaling; it likely requires different configurations or learning strategies to handle problems defined over larger domains effectively.
 
The lower section of the table presents the results for cPIKAN under alternative optimization schemes, including LBFGS and a hybrid ADAM$+$LBFGS approach. Both methods yield substantial improvements in accuracy, reducing the error by over an order of magnitude compared to the ADAM-only version, while maintaining the same per-iteration computational cost. This highlights the potential of optimization strategy in enhancing the training efficiency and precision of physics-informed models.

Figure~\ref{fig:NTK_HelH_RelE} depicts a comparison of three different models, cPIKAN, PINN, and bPIKAN, for solving the Helmholtz equation (Experiment~\ref{Exam.HelmHEqu}). Among the three, cPIKAN demonstrates accurate performance, closely matching the ground truth both qualitatively and quantitatively, with negligible absolute error. PINN can capture the general structure of the solution, including the location of peaks and troughs, but it significantly misestimates their amplitudes, leading to larger absolute errors. In contrast, bPIKAN fails to solve the problem accurately under the given settings, as its prediction deviates substantially from the true solution in both shape and scale.
The plot of relative $\mathcal{L}^2$ error over training epochs shows that cPIKAN achieves the lowest error and exhibits a consistent downward trend, indicating stable convergence. PINN shows slower convergence with a higher final error, and bPIKAN maintains a high relative $\mathcal{L}^2$ error throughout, suggesting poor learning. The loss curves reinforce this behavior: cPIKAN maintains the lowest loss, converging smoothly, while PINN and bPIKAN show higher and less stable losses, with bPIKAN showing clear signs of underfitting or poor training dynamics.

In Fig.~\ref{fig:NTK_HelH}, the evolution of the eigenvalue spectra of the NTK matrix is shown for cPIKAN (top row), PINN (middle row), and bPIKAN (bottom row) throughout the training process. For cPIKAN, the spectra of all NTK-related matrices remain stable and converge smoothly across training iterations. This consistent spectral behavior aligns with the model’s strong performance observed earlier.
In contrast, for PINN, the spectra start to stabilize only after several thousand training steps, and convergence is only observed toward the final epochs. This delayed spectral alignment is consistent with its slower convergence and moderate accuracy.
For bPIKAN, the eigenvalue spectra are highly disordered and do not show clear convergence, indicating instability in training. This chaotic behavior supports the earlier observation that bPIKAN fails to accurately approximate the solution to the Helmholtz equation under the given configuration.

We examine the behavior of the NTK matrix under different optimization strategies for the cPIKAN method, as shown in Figs.~\ref{fig:NTK_HelH_OPT_RelE}-\ref{fig:NTK_HelH_OPT}. The corresponding network architecture used in all experiments is summarized in Table~\ref{Tab.HelmHSetting}.
 Figure~\ref{fig:NTK_HelH_OPT_RelE}  compares the predicted solutions of the Helmholtz equation using three optimizers: ADAM, LBFGS, and a combination of ADAM followed by LBFGS. Visually, the predictions from LBFGS and ADAM$+$LBFGS show excellent agreement with the ground truth, with very low absolute error across the domain. In contrast, ADAM alone yields a noticeably higher absolute error, particularly near the center of the domain. The relative $\mathcal{L}^2$ error plot shows that ADAM+LBFGS achieves the lowest error and fastest convergence, followed by LBFGS. ADAM alone converges more slowly and to a higher final error. This trend is also reflected in the loss curves: both LBFGS and ADAM$+$LBFGS exhibit smooth and steep loss decay, while ADAM shows oscillations and slower reduction.

Figure~\ref{fig:NTK_HelH_OPT} shows the evolution of the eigenvalue spectra of the NTK matrix during the training process for solving the Helmholtz equation, evaluated at different training steps. It compares three optimization strategies: ADAM (left), L-BFGS (center), and hybrid started with ADAM and  followed by LBFGS (right). While all methods lead to changes in the NTK spectrum over time, LBFGS and hybrid approaches demonstrate a more rapid and structured shift in the eigenvalues, especially in the early stages of training. This behavior indicates effective optimization and better conditioning of the NTK matrix, which aligns with their strong performance in terms of both loss and relative $\mathcal{L}^2$ error. Although ADAM shows a smoother and more gradual evolution of the eigenvalues, its final solution is less accurate than the other two methods. This suggests that a well-conditioned NTK spectrum alone does not guarantee high solution accuracy; the choice of optimizer plays a crucial role in how effectively the model explores the solution space.


\clearpage

\subsection{Experiment III:  NTK Behavior in Subdomains (Allen-Cahn Equation)}\label{Exam.AC}
In this example, we investigate the behavior of the NTK when solving the Allen–Cahn equation using the cPIKAN method. We aim to explore whether splitting the domain into smaller temporal subdomains can lead to a more convergent or structured NTK matrix. This could provide valuable insight into the potential of domain decomposition for improving learning stability and efficiency in physics-informed models.\\
We consider the Allen–Cahn equation, which is commonly used to model phase separation in multi-component systems. The equation is given by,
\begin{equation}\label{Eq.AC}
\begin{array}{cc}
     u_{t} - D \, u_{xx} + 5 \left(u^3 - u\right) = 0,& x \in [-6, 6], \, t \in (0, 1], \\[3mm]
     u(-6, t) = u(6, t) = -1, &\quad t \in (0, 1], \\[3mm]
     u(x, 0) = (x/6)^2 \cos\left(\pi \:x/6\right), &\quad x \in [-6, 6],
\end{array}
\end{equation}  
where $u$ is the state variable and $D = 10^{-4}$ is the diffusion coefficient. The nonlinear term $5(u^3 - u)$ governs the local phase dynamics. The initial and boundary conditions are chosen to ensure well-posedness.

\begin{table}[!h]
\centering
\caption{\label{Tab.ACSetting} 
Network configurations, number of trainable parameters $\vert \boldsymbol{\theta}\vert$, number of training data, relative $\mathcal{L}^2$ errors (RE), and per-iteration computational time (in milliseconds) for solving the Allen–Cahn equation (Experiment~\ref{Exam.AC}) using the cPIKAN method. Each row corresponds to a different number of temporal subdomains. The listed settings are applied to each subdomain individually.}
\renewcommand{\arraystretch}{1.2}
\setlength{\tabcolsep}{12pt}
\begin{tabular}{c|ccc|cc}
\toprule
\#  Subdomains & $(N_l, N_n, k)$ & $\vert \boldsymbol{\theta}\vert$& \((N_{\text{r}}, N_{\text{d}})\) & RE & Time \\
\midrule
1 Sub. & (4, 27, 3) & 9072&(8000, 6200) &$5.09 \times 10^{-1}$ & 40 \\
\midrule
2 Subs. & (4, 20, 3) & 5040 &(4000, 3200) & $7.90 \times 10^{-3}$& 20\\
\midrule
4 Subs. & (4, 15, 3)& 2880 & (2000, 1700) &$6.21 \times 10^{-3}$& 20\\
\bottomrule
\end{tabular}
\end{table}

\begin{figure}[!h]
    \centering
    \includegraphics[width=0.9\linewidth]{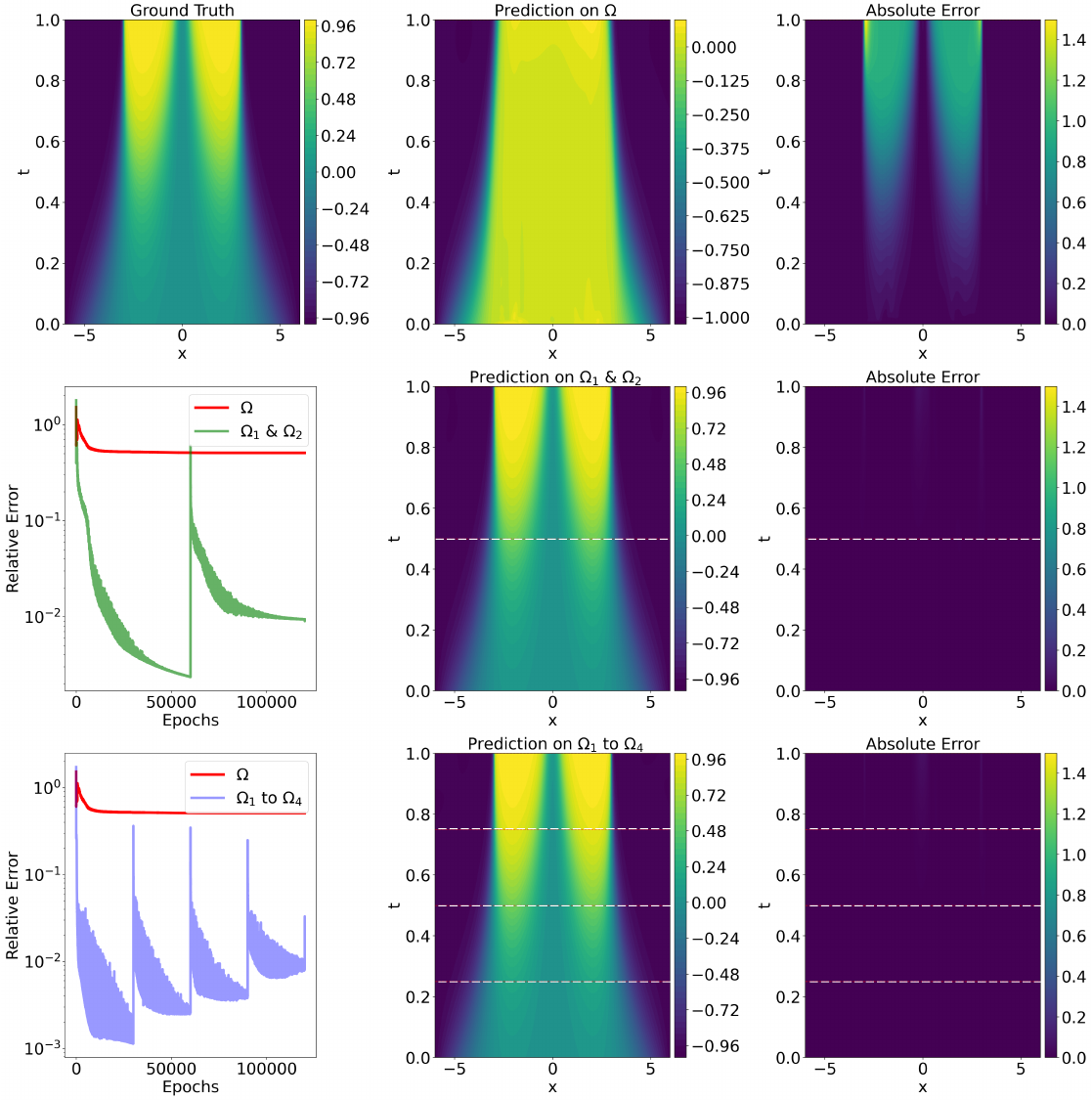}
    \caption{Comparison of the predicted solutions for the Allen–Cahn equation (Experiment~\ref{Exam.AC}) using the cPIKAN method with varying numbers of temporal subdomains. It is shown that increasing the number of subdomains significantly enhances prediction accuracy and convergence. Specifically, the maximum absolute error drops from 1.44 (single domain) to 0.09 and 0.0675 with 2 and 4 subdomains, respectively. The corresponding relative error curves reveal faster and more stable convergence in the multi-subdomain settings, especially evident at the onset of each subdomain. 
}
    \label{fig:NTK_AC_RelE}
\end{figure}

In this experiment, the temporal domain is divided into multiple subdomains to investigate the effect of domain decomposition on training performance using the cPIKAN method. The time interval is split into 1, 2, or 4 equal subdomains. Each subdomain, denoted by $\Omega_i$, is rescaled following the scaling described in \cite{mostajeran2025scaled}, allowing the learning process to be applied consistently within each interval. The predicted solution at the final time of subdomain $\Omega_i$ is used as the initial condition for the next subdomain $\Omega_{i+1}$, and this procedure continues until the final subdomain is reached.

The results for different numbers of subdomains are reported in Table~\ref{Tab.ACSetting}, along with the corresponding network configurations. The networks are designed such that the total number of trainable parameters and training data remains consistent across the entire domain. For example, when the domain is split into 2 subdomains, each subdomain uses 4000 residual points and a network with 5040 parameters. This results in a total of 8000 residual points and 10080 parameters, which is comparable to the case with 1 subdomain (8000 residual points and 9072 parameters).

\begin{figure}[!h]
    \centering
    \includegraphics[width=0.9\linewidth]{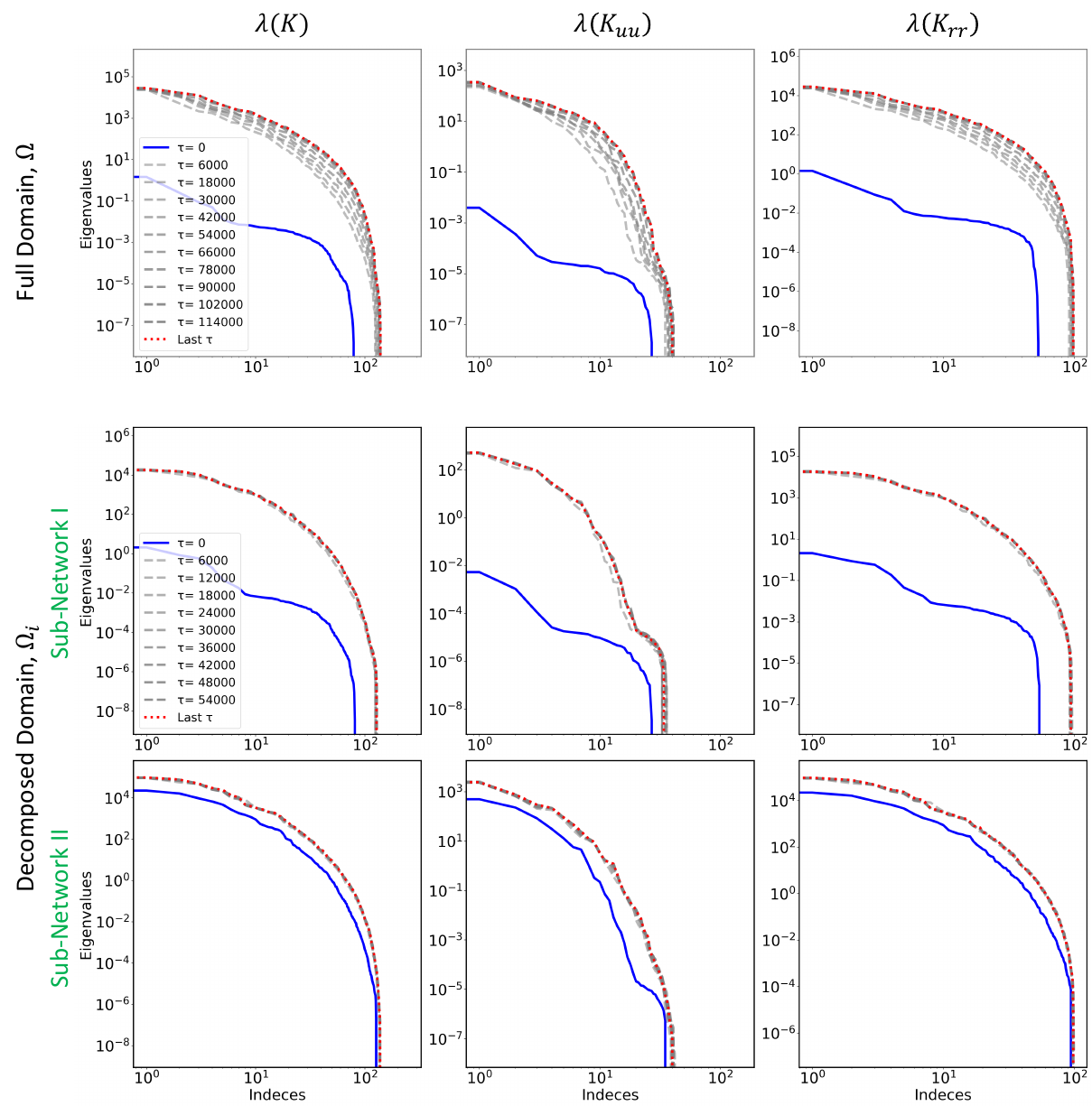}
    \caption{Evolution of the NTK eigenvalue spectra during training of the cPIKAN method for solving the Allen–Cahn equation (Experiment~\ref{Exam.AC}). The comparisons show that increasing the number of temporal subdomains leads to faster spectral convergence and better conditioning of key NTK blocks. This improved behavior reflects more efficient learning dynamics and reduced spectral bias. }
    \label{fig:NTK_AC_Seq2}
\end{figure}

As shown in the table, using 2 or 4 subdomains significantly reduces the relative $\mathcal{L}^2$ error compared to training a single network over the full domain. In particular, the error drops from $5.09 \times 10^{-1}$ (1 subdomain) to $7.90 \times 10^{-3}$ (2 subdomains) and further to $6.21 \times 10^{-3}$ (4 subdomains). Additionally, the required runtime per iteration is reduced from 40 milliseconds in the single-domain case to 20 milliseconds when using subdomain decomposition, demonstrating both improved accuracy and computational efficiency. In Fig.~\ref{fig:NTK_AC_RelE}, we evaluate the performance of the proposed cPIKAN method when applied to the Allen–Cahn equation using different numbers of temporal subdomains. The predicted solutions and corresponding absolute errors demonstrate that increasing the number of subdomains significantly improves accuracy. For example, the maximum absolute error decreases from  1.44  in the single-domain case to  0.09  with 2 subdomains and further to  0.0675  with 4 subdomains. This improvement is also reflected in the relative error plots (left column), where the error curves for the multi-subdomain cases show a consistent downward trend, reaching lower magnitudes than the single-domain counterpart. Notably, in the 2- and 4-subdomain settings, the relative error sharply drops at the start of each subdomain, indicating efficient learning within localized temporal regions. These observations confirm that temporal domain decomposition not only enhances convergence but also results in a more accurate and stable solution across the entire spatiotemporal domain.

\begin{figure}[!h]
    \centering
    \includegraphics[width=0.9\linewidth]{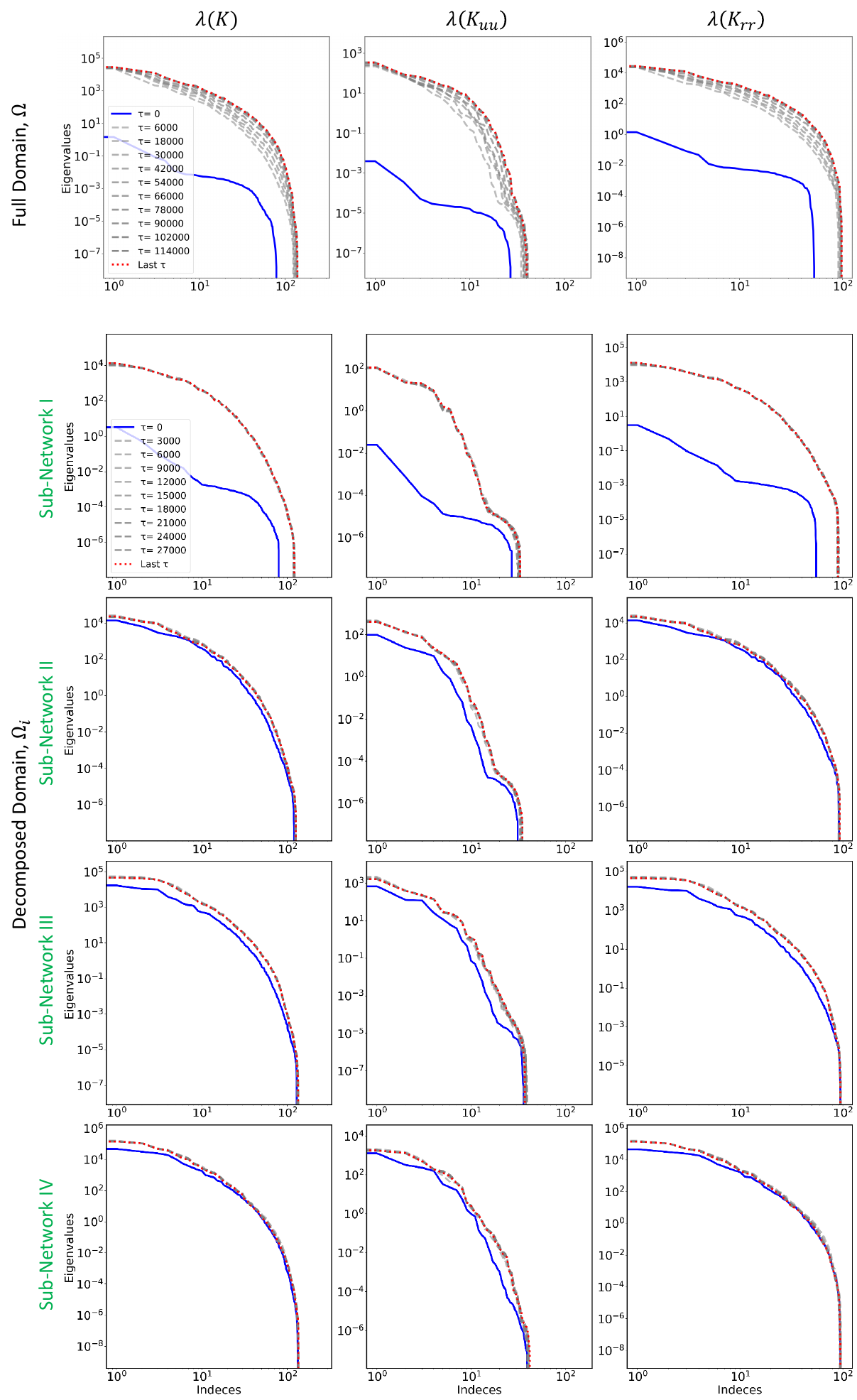}
    \caption{ Evolution of the NTK eigenvalue spectra during training of the cPIKAN method for solving the Allen–Cahn equation (Experiment~\ref{Exam.AC}). The comparison shows that increasing the number of temporal subdomains leads to faster spectral convergence and better conditioning of key NTK blocks. This improved behavior reflects more efficient learning dynamics and reduced spectral bias.} 
    \label{fig:NTK_AC_Seq4}
\end{figure}

Figures \ref{fig:NTK_AC_Seq2} and \ref{fig:NTK_AC_Seq4} illustrate the evolution of the NTK eigenvalue spectra during training of the cPIKAN method for the Allen–Cahn equation. As the number of temporal subdomains increases, the NTK matrices become increasingly well-conditioned and converge more rapidly. This behavior highlights how domain decomposition mitigates spectral bias and enhances the learning dynamics of cPIKAN. The strategy reduces the complexity of the target function in each sub-network, leading to more stable training and improved predictive accuracy. This confirms that spectral bias is a critical factor limiting performance in the full domain case, and domain decomposition offers a practical remedy, especially in diffusion-dominated regimes like Allen–Cahn.


\clearpage
\subsection{Experiment IV: High-Order Dynamics and Enriched NTK Structure  (forced vibration Equation)}\label{Exam:Vib}

We now consider a forced vibration problem governed by the Euler–Bernoulli beam equation with damping. This example is used to highlight the structural complexity of the associated NTK matrix, which differs from previous cases due to the inclusion of higher-order temporal and spatial derivatives.
The governing equation is given by,
\begin{equation}
D_f \frac{\partial^4 u(x,t)}{\partial x^4} + c_d \frac{\partial u(x,t)}{\partial t} + \rho_l \frac{\partial^2 u(x,t)}{\partial t^2} = p(x,t), \quad (x,t) \in (0,l)\times (0,T],
\end{equation}
where $D_f$ is the flexural stiffness, $c_d$ is the damping coefficient, $\rho_l$ is the mass per unit length, $p$ is the external excitation force, and $u(x,t)$ is the displacement field.
This equation is subject to the following boundary and initial conditions,
\begin{equation}
    \begin{array}{ccc}
    \begin{cases}
u(0,t) = u(l,t) = 0, \\
\displaystyle\frac{\partial^2 u}{\partial x^2}(0,t) = \displaystyle\frac{\partial^2 u}{\partial x^2}(l,t) = 0,
\end{cases}
\quad t \in [0,T],     &   
\text{and}
&
\begin{cases}
u(x,0) = 0, \\
\displaystyle\frac{\partial u}{\partial t}(x,0) = 0,
\end{cases}
\quad x \in (0,l).
    \end{array}
\end{equation}

The parameters of the beam used in this example are as follows: $l = 1$ m, $\rho_l = 1$ kg/m, $D_f = 2$ N.m$^2$, $c_d = 3$ N.s/m$^2$. The excitation force is $p(x,t) = P \sin(\pi x/l)\cos(2 \pi f t)$, with an amplitude of $P=0.1$ N/m and a frequency of $f=2.7$ Hz. 

In contrast to the previous examples, the NTK matrix in this case includes additional diagonal blocks corresponding to the temporal derivative $u_t$ and the second-order spatial derivative $u_{xx}$, in addition to the standard terms $K_{uu}$ and $K_{rr}$. 
Within the physics-informed learning framework, we define the stacked network output $\boldsymbol{\psi}(\tau)$, the corresponding target vector $\boldsymbol{\mathcal{G}}$, and the associated NTK matrix $\boldsymbol{K}_{\text{ntk}}(\tau)$ in Eq.~\eqref{Eq:NTK100} as follows,
\begin{equation}\label{Eq:VibNTK}
    \begin{array}{ccc}
        \boldsymbol{\psi}(\tau) = \begin{bmatrix} 
    u(\boldsymbol{x}^d_i; \boldsymbol{\theta}(\tau)) \\
    u_t(\boldsymbol{x}^0_i; \boldsymbol{\theta}(\tau)) \\
    u_{xx}(\boldsymbol{x}^b_i; \boldsymbol{\theta}(\tau)) \\
    \mathcal{N}[u](\boldsymbol{x}^r_i; \boldsymbol{\theta}(\tau))
\end{bmatrix},
&  
\boldsymbol{\mathcal{G}} = \begin{bmatrix} 
    \boldsymbol{0} \\
    \boldsymbol{0} \\
    \boldsymbol{0} \\
    \boldsymbol{p}(\boldsymbol{x}^r_i)
\end{bmatrix},
&
\boldsymbol{K}_{\text{ntk}}(\tau) = 
\begin{bmatrix}
    K_{u,u} & K_{u,u_t} & K_{u,u_{xx}} & K_{u,r} \\
    K_{u_t,u} & K_{u_t,u_t} & K_{u_t,u_{xx}} & K_{u_t,r} \\
    K_{u_{xx},u} & K_{u_{xx},u_t} & K_{u_{xx},u_{xx}} & K_{u_{xx},r} \\
    K_{r,u} & K_{r,u_t} & K_{r,u_{xx}} & K_{r,r}
\end{bmatrix},
    \end{array}
\end{equation}
in which the  diagonal blocks are computed as inner products of gradients with respect to the model parameters,
\begin{equation}
\begin{array}{ll}
(K_{u,u})_{i,j} = \displaystyle{\left\langle \frac{\partial u(\boldsymbol{x}^d_i; \boldsymbol{\theta}(\tau))}{\partial \boldsymbol{\theta}}, \frac{\partial u(\boldsymbol{x}^d_j; \boldsymbol{\theta}(\tau))}{\partial \boldsymbol{\theta}} \right\rangle}, &
 (K_{u_{xx},u_{xx}})_{i,j} = \displaystyle{\left\langle \frac{\partial u_{xx}(\boldsymbol{x}^b_i; \boldsymbol{\theta}(\tau))}{\partial \boldsymbol{\theta}}, \frac{\partial u_{xx}(\boldsymbol{x}^b_j; \boldsymbol{\theta}(\tau))}{\partial \boldsymbol{\theta}} \right\rangle},\\[4mm] 
(K_{u_t,u_t})_{i,j} = \displaystyle{\left\langle \frac{\partial u_t(\boldsymbol{x}^0_i; \boldsymbol{\theta}(\tau))}{\partial \boldsymbol{\theta}}, \frac{\partial u_t(\boldsymbol{x}^0_j; \boldsymbol{\theta}(\tau))}{\partial \boldsymbol{\theta}} \right\rangle},&
(K_{r,r})_{i,j} = \displaystyle{\left\langle \frac{\partial \mathcal{N}[u](\boldsymbol{x}^r_i; \boldsymbol{\theta}(\tau))}{\partial \boldsymbol{\theta}}, \frac{\partial \mathcal{N}[u](\boldsymbol{x}^r_j; \boldsymbol{\theta}(\tau))}{\partial \boldsymbol{\theta}} \right\rangle}.
\end{array}
\end{equation}
leading to a larger and more structured NTK, capturing richer dynamics of the system and posing a more complex challenge for the learning process.

\begin{table}[!h]
\centering
\caption{\label{Tab.VibSetting} 
Network configurations, number of trainable parameters $\vert \boldsymbol{\theta}\vert$, number of residual training data, number of epochs in training each subnetwork, relative $\mathcal{L}^2$ errors (RE), and per-iteration computational time (in milliseconds) for solving the forced vibration equation (Experiment~\ref{Exam:Vib}) when $T=10$ (s) using the (a) cPIKAN and (b) PINN method. Each row corresponds to a different number of temporal subdomains. The listed settings are applied to each subdomain individually.
Each model is trained with 400 points for initial conditions and 400 for boundary conditions.
}
\renewcommand{\arraystretch}{1.2}
\setlength{\tabcolsep}{12pt}
\begin{tabular}{c|cccc|cc}
\toprule
\multicolumn{7}{l}{(a) cPIKAN method} \\
\midrule
\#  Subdomains & $(N_l, N_n, k)$ & $\vert \boldsymbol{\theta}\vert$& \(N_{\text{r}}\)&\# epochs & RE & Time \\
\midrule
1 Sub. & (4, 47, 5) & 40608&40000&$10^6$ &$7.64 \times 10^{0}$ & 660 \\
\midrule
2 Subs. & (4, 34, 5) & 21420 &20000& $5\times 10^5$& $9.57 \times 10^{0}$& 240\\
\midrule
4 Subs. & (4, 24, 5)& 10800 & 10000& $25\times 10^4$ &$3.55 \times 10^{-2}$& 83\\
\midrule
8 Subs. & (4, 17, 5)& 5508 & 5000& $125\times 10^3$ &$7.06 \times 10^{-3}$& 74\\
\midrule
\midrule
\multicolumn{7}{l}{(b) PINN method} \\
\midrule
\#  Subdomains & $(N_l, N_n)$ & $\vert \boldsymbol{\theta}\vert$& \(N_{\text{r}}\)&\# epochs & RE & Time \\
\midrule
8 Sub. & (4, 40) & 5080&5000& $125\times 10^3$ &$7.79 \times 10^{-1}$ & 16 \\
\bottomrule
\end{tabular}
\end{table}
%

Similar to the previous experiment (Experiment~\ref{Exam.AC}), the temporal domain in this experiment is also divided into multiple subdomains, each denoted by $\Omega_i$ and individually rescaled. The solution at the final time of $\Omega_i$ is used as the initial condition for $\Omega_{i+1}$, enabling sequential training across subdomains.
Table~\ref{Tab.VibSetting} summarizes the performance of the cPIKAN and PINN methods in solving the forced vibration equation over a time interval $[0, T]$ with final time $T = 10$ seconds, using different numbers of temporal subdomains. The settings for each subnetwork are chosen such that the total number of parameters, training data, and training iterations across the entire domain remain consistent.
For the cPIKAN method, increasing the number of subdomains leads to a significant improvement in accuracy and a notable reduction in training time. 
According to the results, the cPIKAN method performs poorly when using only 1 (i.e., full domain) or 2 subdomains, as indicated by the large relative errors. In these cases, the network fails to provide an accurate approximation of the solution, despite requiring a substantial amount of training time. The performance improves significantly when the domain is divided into 4 subdomains, leading to both lower error and reduced computational time. 
Further decomposition into 8 subdomains yields additional benefits. Specifically, the relative error decreases by approximately 80\% compared to the 4-subdomain case, and the total training time is reduced by about 11\%. This demonstrates that increasing the number of subdomains not only enhances accuracy but also improves training efficiency.
This highlights the advantage of temporal decomposition in enhancing both accuracy and time. 
When using 8 subdomains, the cPIKAN method delivers a solution that is approximately 99\% more accurate than the standard PINN approach. Although cPIKAN requires more training time, this additional cost results in a dramatic improvement in accuracy, making it a highly effective choice for problems where precision is critical. These results demonstrate that cPIKAN successfully achieve high-resolution solutions that PINNs cannot match, even under the same training settings per subdomain.

\begin{figure}[!h]
    \centering
    \includegraphics[width=1.0\linewidth]{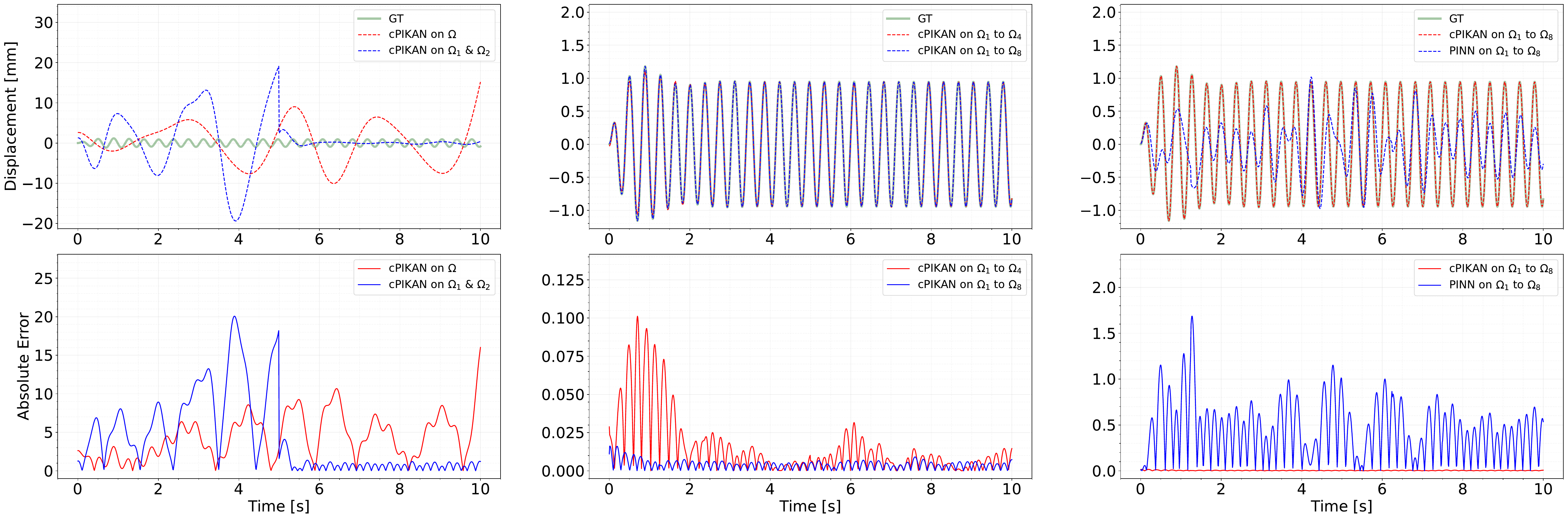}
    \caption{Comparison of predicted solutions for the forced vibration equation (Experiment~\ref{Exam:Vib}) using cPIKAN with varying numbers of temporal subdomains, demonstrating that increasing the number of subdomains dramatically improves accuracy. While configurations with 1 or 2 subdomains yield large errors (up to 20.07), using 4 and 8 subdomains reduces the maximum absolute error to 0.102 and 0.016, respectively, an 84\% improvement. Moreover, cPIKAN with 8 subdomains outperforms PINN with the same decomposition by approximately 98\%, confirming that the temporal domain decomposition strategy enhances both stability and predictive precision, especially in problems involving oscillatory dynamics.
    }
    \label{fig:Vib_RelE}
\end{figure}

Figure~\ref{fig:Vib_RelE}  compares the performance of the cPIKAN method with 1, 2, 4, and 8 temporal subdomains, as well as the PINN method with 8 subdomains. In terms of accuracy, the maximum absolute error for cPIKAN with 1 and 2 subdomains is very large, 16.01 and 20.07, respectively, indicating that the network fails to provide a reliable solution in these cases. However, the error drops significantly when the number of subdomains increases. For 4 subdomains, the maximum error decreases to 0.102, and with 8 subdomains, it reaches as low as 0.016. This represents an 84\% improvement in accuracy from 4 to 8 subdomains within the cPIKAN framework. Compared to PINN with 8 subdomains, which yields a maximum error of 0.779, the 8-subdomain cPIKAN solution is approximately 98\% more accurate, highlighting the benefit of the domain decomposition strategy.
%

\begin{figure}[!h]
    \centering
    \includegraphics[width=1.0\linewidth]{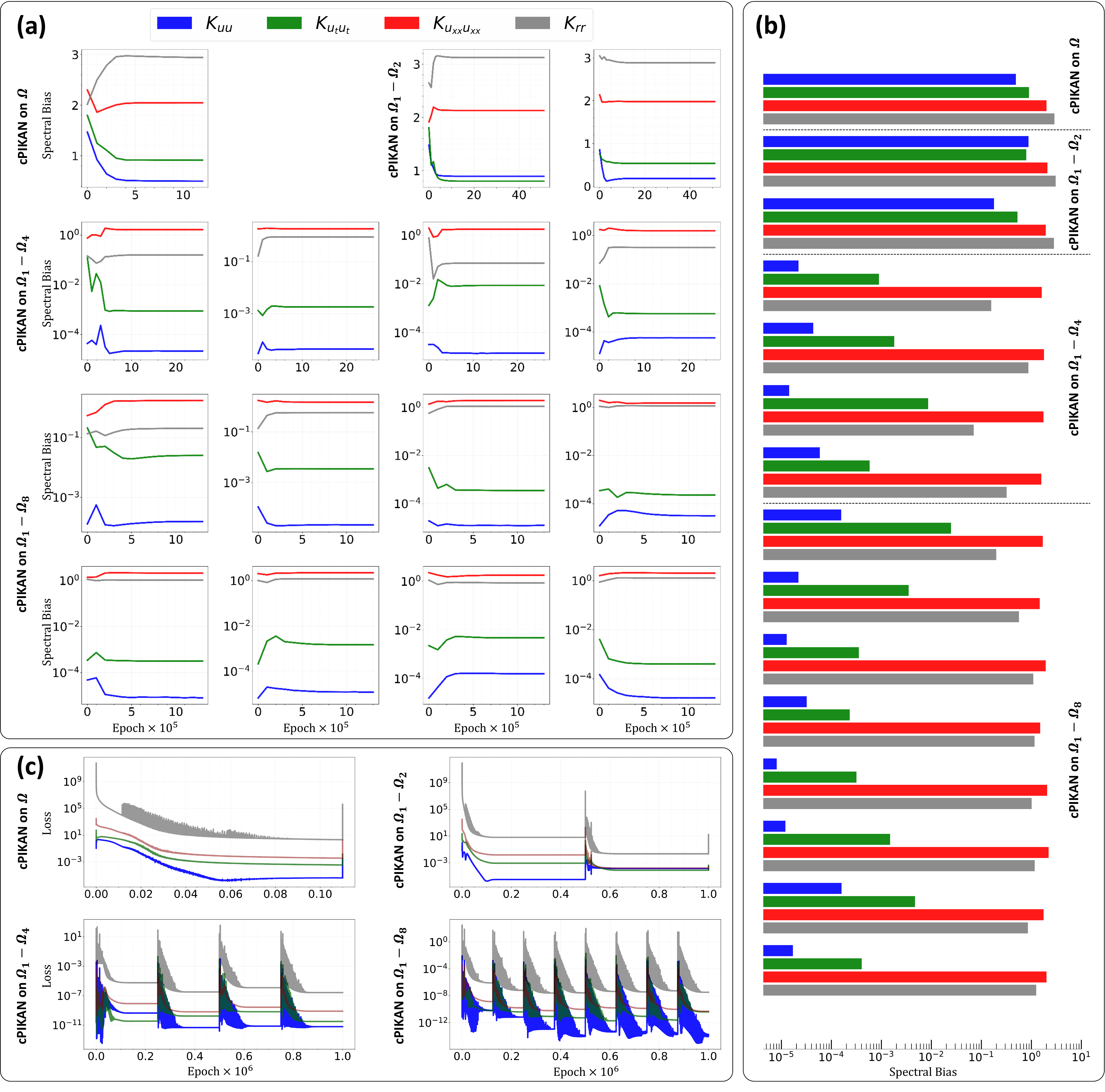}
    \caption{ Comparison of the training dynamics and spectral entropy behavior of the cPIKAN method for the forced vibration equation (Experiment~\ref{Exam:Vib}) under varying temporal subdomain configurations. Panels (a) and (b) shows that increasing the number of subdomains consistently reduces the spectral entropy of key NTK components, particularly $K_{uu}$ and $K_{u_t u_t}$, indicating lower kernel complexity and more structured learning dynamics. This reduction becomes more pronounced as the number of subdomains increases from 1 to 8. Additionally, the training loss curves shown in Panel (c) reveal faster and more stable convergence for configurations with more subdomains. These findings demonstrate that temporal decomposition not only improves training efficiency but also enhances the generalization capacity of cPIKAN by promoting simpler and more focused representations in the neural tangent kernel.}
    \label{fig:VibSEcPIKAN}
\end{figure}

Rather than illustrating the progression of the diverse eigenvalues of the NTK matrices, we introduce an alternative metric dubbed as the spectral entropy of the NTK matrix, calculated as,
\begin{equation}\label{Eq.shanon}
    \text{Spectral Entropy} = -\sum_i \frac{|\lambda_i|}{\sum_j |\lambda_j|} \log \left( \frac{|\lambda_i|}{\sum_j |\lambda_j|} \right),
\end{equation}
where $\lambda_i$ are the eigenvalues of the NTK matrix.
We adopt the concept of spectral entropy from Shannon’s information theory \cite{shannon1948mathematical}, introduced in the 1940s to quantify uncertainty and information content in communication systems. Shannon’s information theory defines entropy as a measure of unpredictability or information content in a signal, which later became a cornerstone in data compression \cite{cover1999elements, ornstein1993entropy, balakrishnan2007relationship} and machine learning \cite{murphy2012machine,sepulveda2024applications} and leveraged as cross-entropy in classification models using deep learning \cite{goodfellow2016deep, bishop2006pattern}.  
In Eq.~\eqref{Eq.shanon}, by applying Shannon’s information theory to the normalized eigenvalues, we compute the spectral entropy that captures the transient and convergence behavior of the NTK spectrum. This helps evaluate how well the network approximates the target function across localized regions of the domain.
Based on Eq.\eqref{Eq.shanon}, a low spectral entropy indicates that the network’s learning dynamics are focused along a few dominant directions, while a high entropy implies more distributed and less structured learning behavior.

Figure~\ref{fig:VibSEcPIKAN} presents a comprehensive analysis of the training behavior and kernel complexity of the cPIKAN method when applied to the forced vibration problem under different subdomain configurations. Panel (a) illustrates the evolution of spectral entropy for various NTK components throughout training, showing how these components converge over time. The results indicate that dividing the domain into subdomains generally leads to a noticeable reduction in spectral entropy for most NTK components, suggesting a simplification of the learning dynamics and more efficient training in structured domains. Panel (b) shows the final spectral entropy values after training, clearly revealing that entropy decreases as the number of subdomains increases, especially for the $K_{uu}$ and $K_{u_t u_t}$ components. This supports the idea that subdomain decomposition helps to reduce kernel complexity and may enhance training stability. Finally, panel (c) presents the training loss curves, which confirm consistent convergence across all configurations. Notably, as the number of subdomains increases, the training process becomes slightly faster and more stable, with the 8-subdomain configuration showing the most efficient convergence behavior.
\begin{figure}[!h]
    \centering
    \includegraphics[width=0.84\linewidth]{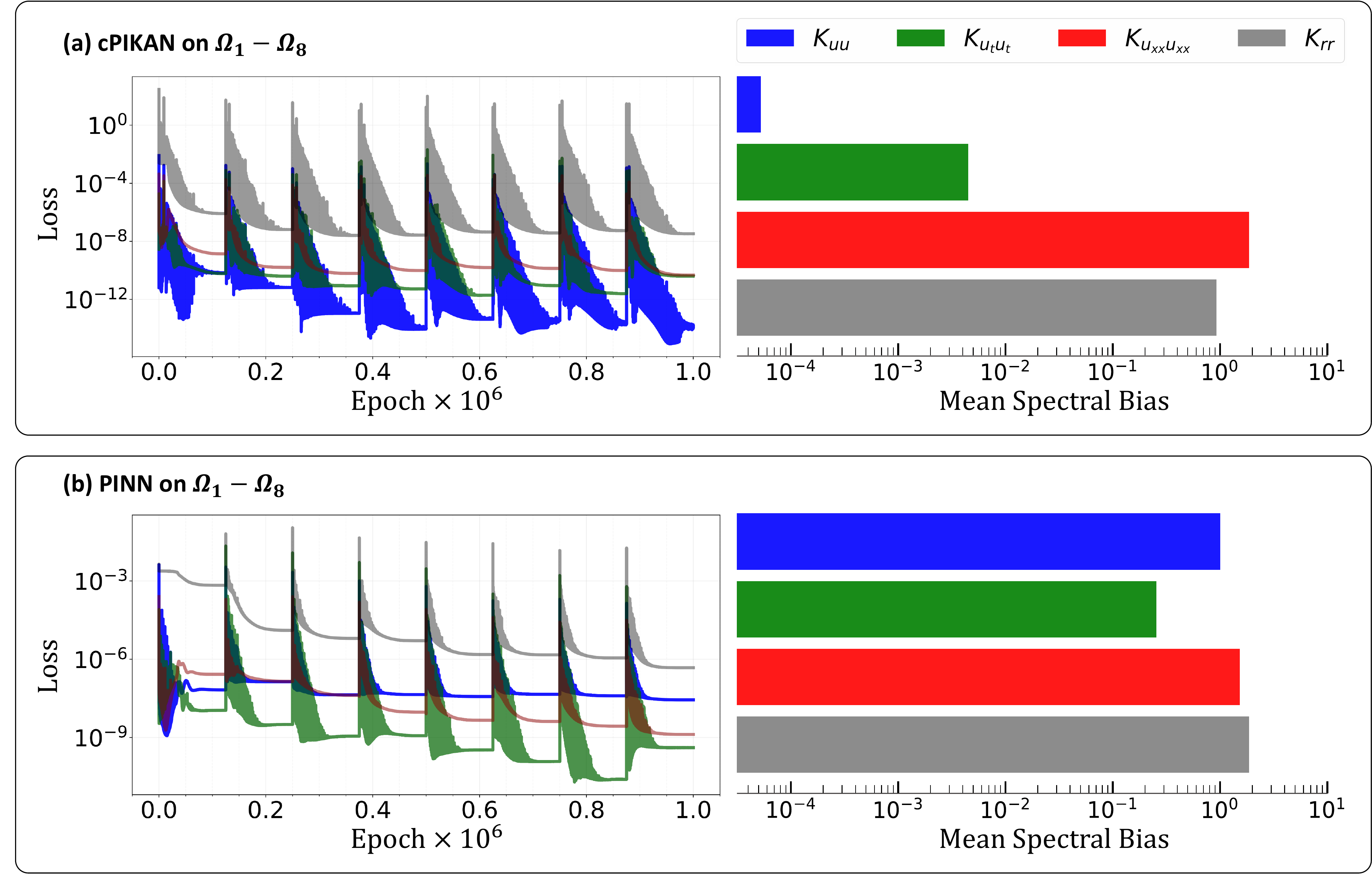}
    \caption{Comparison of training dynamics and spectral bias between (a) cPIKAN and (b) PINN for the forced vibration problem (Experiment~\ref{Exam:Vib}) with 8 temporal subdomains. The comparisons show that cPIKAN achieves faster and more stable convergence, as evidenced by sharper and smoother loss reduction. Additionally, cPIKAN exhibits consistently lower mean spectral entropy across key NTK components, especially $K_{uu}$ and $K_{u_t u_t}$, indicating reduced kernel complexity and enhanced representation of solution features. These findings highlight the advantage of cPIKAN in capturing the underlying physics more efficiently and accurately than PINN, particularly in problems with oscillatory or high-frequency behavior.    
    }
    \label{fig:VibSEPINN}
\end{figure}

Figure~\ref{fig:VibSEPINN} compares the training performance and kernel behavior of the cPIKAN and PINN methods for the forced vibration problem, both using the same division into 8 temporal subdomains. The loss plots on the left show that cPIKAN converges more quickly and consistently than PINN, with sharper drops in loss values and more regular training dynamics. On the right, the mean spectral entropy plots indicate that cPIKAN has lower  mean spectral entropy for the key solution-related NTK components, especially $K_{uu}$ and $K_{u_t u_t}$, suggesting that it better captures high-frequency features of the solution. These observations highlight cPIKAN’s improved ability to learn the underlying physics of the problem more efficiently and accurately compared to PINN.

\section{ Future Outlook}





In the future, we envision three key directions to expand this work. 
First, a deeper theoretical understanding is needed to determine how the NTK of cPIKANs behaves with respect to the model’s complexity, specifically, the polynomial degree and the number of layers. 
As shown in \cite{seleznova2022analyzing}, for MLPs initialized in a regime that leads to gradient explosion across layers, the NTK does not follow classical theory. Instead of remaining stable, the kernel appears random at initialization and evolves significantly during training. This challenges the traditional assumption that the NTK remains nearly constant throughout training in sufficiently wide networks.
Investigating whether similar phenomena occur in cKANs and how the Chebyshev structure influences such dynamics remains an open question.   Second, in this work, we examined the eigenvalue distribution and spectral entropy of the NTK matrix to gain insight into the learning dynamics of cPIKAN. While these metrics offer a useful first look into the structure and evolution of the kernel, they do not fully explain how the NTK influences model accuracy or generalization. A promising direction for future research is to develop additional theoretical measures that more directly link the NTK spectrum to the model’s prediction error. For example, analyzing how the alignment between the NTK and the target function changes during training, or investigating how the energy concentration in the leading eigenmodes relates to final performance, could provide a deeper understanding of the learning process. Such studies may lead to the development of new error estimators or performance predictors derived from NTK properties, offering a stronger theoretical basis for architecture design and training decisions. Third, we observed that as the number of boundary and initial conditions increases, such as in the forced vibration problem, the NTK matrix grows in size, leading to significant computational overhead. This issue becomes more pronounced in problems involving coupled PDE systems or high-dimensional physics-based models. To make the NTK-based analysis tractable in such cases, future work should explore matrix decomposition techniques or develop alternative learning metrics that capture the essence of learning dynamics at a lower computational cost.

\section{ Conclusions}

This study investigated the training behavior of Chebyshev-based physics-informed Kolmogorov–Arnold Networks  (cPIKANs) through the framework of Neural Tangent Kernel (NTK) theory. We first derived and analyzed the NTK for cKANs in the supervised learning setting, providing a theoretical foundation for understanding how the kernel structure governs convergence dynamics. Building on this, we extended the analysis to cPIKANs using numerical experiments across a variety of PDEs, including the diffusion equation, Helmholtz equation, Allen–Cahn equation, and a forced vibration problem. 
Through these case studies, we observed that the NTK matrices associated with  cPIKANs exhibit structured and tractable behavior during training. The spectral analysis revealed consistent patterns, such as eigenvalue concentration and reduced entropy, that directly correlate with faster convergence and improved generalization. These findings demonstrate that the NTK framework not only explains the empirical advantages of cPIKANs, but also provides predictive insight into their learning dynamics across different PDE settings.

In the one-dimensional diffusion equation, we compared cPIKAN with two PINN baselines and found that cPIKAN achieved up to 7 times higher accuracy and significantly faster convergence. This performance gain was associated with a more informative and well-conditioned NTK eigenvalue spectrum, which exhibited slower decay and greater stability throughout training. These spectral characteristics enhanced the tractability of the learning dynamics and allowed the model to retain and propagate useful gradient information more effectively.
In the two-dimensional Helmholtz equation, we compared cPIKAN, PINN, and bPIKAN, and found that cPIKAN achieved an error approximately three times smaller than PINN and five times smaller than bPIKAN. NTK analysis showed that cPIKAN maintained a more stable and better-conditioned eigenvalue spectrum, resulting in improved convergence and accuracy. Additionally, using LBFGS or a hybrid ADAM+LBFGS optimizer, the error was reduced by nearly an order of magnitude compared to ADAM alone, highlighting the important influence of optimization strategy on NTK dynamics and model performance.
The Allen-Cahn experiment shows that splitting the time domain into 2 or 4 subdomains improves the NTK conditioning and convergence, reducing the relative error by over 98\%  and the maximum absolute error by over 95\%, compared to training on the full domain.
In the final experiment, we considered a forced vibration problem governed by a high-order PDE, which led to a larger and more structured NTK matrix due to the presence  of higher order derivative terms.
Results show that cPIKAN with 8 temporal subdomains achieves over 98\% lower maximum error compared to PINN, while also reducing spectral entropy of key NTK blocks by a factor of three, indicating more focused and efficient learning dynamics. This highlights the dual benefit of temporal decomposition in both reducing NTK complexity and improving model accuracy and convergence stability.

\section{Acknowledgements}
S.A.F. acknowledges the support by the U.S. Department of Energy’s Office of Environmental Management  (award no.: DE-EM0005314).

\section{Conflict of Interest}
The authors declare no conflict of interest.

\def\mybibdoicolor{\color{black}}
\newcommand*{\doi}[1]{\href{\detokenize{#1}} {\raggedright\mybibdoicolor{DOI: \detokenize{#1}}}}

\bibliographystyle{elsarticle-num}
\bibliography{references.bib}

\newpage
\appendix
\renewcommand{\thefigure}{A.\arabic{figure}}
\setcounter{figure}{0}

\section{Proof of Theorem \ref{Theorem:NTKcKAN}}\label{App.Proof1}

The Neural Tangent Kernel between two inputs $\boldsymbol{x}$ and $\boldsymbol{x}'$ is defined as the inner product of the gradients of the network output with respect to the parameters,
\begin{equation}\label{Eq:K_NTK_xx}
\begin{array}{cl}
    \boldsymbol{K}_{\text{ntk}}(\boldsymbol{x}, \boldsymbol{x}') & = \displaystyle{\left\langle \frac{\partial f(\boldsymbol{x}; \boldsymbol{\theta}(0))}{\partial \boldsymbol{\theta}}, \frac{\partial f(\boldsymbol{x}'; \boldsymbol{\theta}(0))}{\partial \boldsymbol{\theta}} \right\rangle} \\[4mm]
     & = \displaystyle{\sum_{i,j,n} \frac{\partial f(\boldsymbol{x}; \boldsymbol{\theta}(0))}{\partial w_{i,j,n}^{(1)}} \cdot \frac{\partial f(\boldsymbol{x}'; \boldsymbol{\theta}(0))}{\partial w_{i,j,n}^{(1)}} + \sum_{j,n} \frac{\partial f(\boldsymbol{x}; \boldsymbol{\theta}(0))}{\partial w_{j,1,n}^{(2)}} \cdot \frac{\partial f(\boldsymbol{x}'; \boldsymbol{\theta}(0))}{\partial w_{j,1,n}^{(2)}}}.
\end{array}
\end{equation}
Now, we simplify the NTK expression statistically, using the fact that the coefficients are i.i.d. standard normal,
\begin{equation}
 w_{i,j,n}^{(1)},\; w_{j,1,n}^{(2)} \sim \mathcal{N}(0, 1).
\end{equation}
We simplify the expected NTK,
\begin{equation}
    \mathbb{E}[\boldsymbol{K}_{\text{ntk}}(\boldsymbol{x}, \boldsymbol{x}')] = \mathbb{E}\left[K^{(1)}(\boldsymbol{x}, \boldsymbol{x}')\right] + \mathbb{E}\left[K^{(2)}(\boldsymbol{x}, \boldsymbol{x}')\right].
\end{equation}
where,
\begin{equation}
    K^{(1)}(\boldsymbol{x}, \boldsymbol{x}') = \displaystyle{\sum_{i,j,n} \frac{\partial f(\boldsymbol{x}; \boldsymbol{\theta}(0))}{\partial w_{i,j,n}^{(1)}} \cdot \frac{\partial f(\boldsymbol{x}'; \boldsymbol{\theta}(0))}{\partial w_{i,j,n}^{(1)}}}, 
    \quad
    K^{(2)}(\boldsymbol{x}, \boldsymbol{x}') = \displaystyle{\sum_{j,n} \frac{\partial f(\boldsymbol{x}; \boldsymbol{\theta}(0))}{\partial w_{j,1,n}^{(2)}} \cdot \frac{\partial f(\boldsymbol{x}'; \boldsymbol{\theta}(0))}{\partial w_{j,1,n}^{(2)}}}.
\end{equation}
This helps us understand the average behavior of the kernel at initialization.

\noindent
\textbf{Step I:} 
In this step, we consider the expectation of the second term,
\begin{equation}
\mathbb{E}[K^{(2)}(\boldsymbol{x}, \boldsymbol{x}')] = N \cdot \sum_{n=0}^k \mathbb{E} \left[ T_n(z_j(\boldsymbol{x})) \cdot T_n(z_j(\boldsymbol{x}')) \right],
\end{equation}
where $z_j(\boldsymbol{x}) = \tanh(h_j(\boldsymbol{x}))$ and $z_j(\boldsymbol{x}') = \tanh(h_j(\boldsymbol{x}'))$ are jointly distributed through the bivariate Gaussian pair $(h_j(\boldsymbol{x}), h_j(\boldsymbol{x}'))$ with zero mean and variances $\sigma_j^2(\boldsymbol{x})$ and $\sigma_j^2(\boldsymbol{x}')$, as defined previously in Eq.~\eqref{Eq:VarianceH}. The covariance between $h_j(\boldsymbol{x})$ and $h_j(\boldsymbol{x}')$ is given by,
\begin{equation}\label{Eq:CovH}
\mathrm{Cov}[h_j(\boldsymbol{x}), h_j(\boldsymbol{x}')] = \sum_{i=1}^d \sum_{n=0}^k T_n(\tanh(x_i)) \cdot T_n(\tanh(x'_i)) =: \rho_j(\boldsymbol{x},\boldsymbol{x}').
\end{equation}
Defining,
\begin{equation}
C_n(\boldsymbol{x}, \boldsymbol{x}') := \mathbb{E}\left[ T_n(\tanh(h)) \cdot T_n(\tanh(h')) \right],
\end{equation}
for $(h, h') \sim \mathcal{N}\left(0, \begin{bmatrix} \sigma^2(\boldsymbol{x}) & \rho(\boldsymbol{x},\boldsymbol{x}') \\ \rho(\boldsymbol{x},\boldsymbol{x}') & \sigma^2(\boldsymbol{x}') \end{bmatrix} \right)$, we obtain the compact expression,
\begin{equation}\label{Eq:StepI}
\mathbb{E}[K^{(2)}(\boldsymbol{x}, \boldsymbol{x}')] = N \cdot \sum_{n=0}^k C_n(\boldsymbol{x}, \boldsymbol{x}').
\end{equation}
Since Chebyshev polynomials are bounded on the interval $(-1, 1)$, and the $\tanh$ function maps Gaussian variables into this range, each term $C_n(\boldsymbol{x}, \boldsymbol{x}')$ remains finite.

\noindent
\textbf{Step II:} 
In the second step, we compute the expectation of the first-term kernel $\mathbb{E}[K^{(1)}(\boldsymbol{x}, \boldsymbol{x}')]$. 
Using Eq.~\eqref{Eq:firstdiff}, the first-term kernel can be written as,
\begin{equation}
K^{(1)}(\boldsymbol{x}, \boldsymbol{x}') = \sum_{i,j,n} A_{i,j,n}(\boldsymbol{x}) \cdot A_{i,j,n}(\boldsymbol{x}'),
\end{equation}
where,
\begin{equation}
A_{i,j,n}(\boldsymbol{x}) = T_n(\tilde{x}_i) \cdot \psi(\boldsymbol{x}) \cdot \sum_{m=0}^k w_{j,1,m}^{(2)} \cdot T_m'(\tanh(h_j(\boldsymbol{x}))),
\end{equation}
and $\psi(\boldsymbol{x}) = (1 - \tanh^2(h_j(\boldsymbol{x})))$.
Taking expectation with respect to the second-layer coefficients $w_{j,1,m}^{(2)} \sim \mathcal{N}(0,1)$, and noting that for fixed $h_j(\boldsymbol{x})$, the sum becomes a linear combination of independent Gaussian variables, we obtain a zero mean and the following variance,
\begin{equation}
\sum_{m=0}^{k} \left( T_m'(\tanh(h_j(\boldsymbol{x}))) \right)^2.
\end{equation}
This leads to the following expectation,
\begin{equation}
    \begin{array}{l}
      \mathbb{E}_{w^{(2)}}[A_{i,j,n}(\boldsymbol{x}) \cdot A_{i,j,n}(\boldsymbol{x}')] =  \\  T_n(\tilde{x}_i) \cdot T_n(\tilde{x}'_i) \cdot 
          \mathbb{E}_{h_j(\boldsymbol{x}), h_j(\boldsymbol{x}')} \Bigg[ \psi(\boldsymbol{x}) \cdot \psi(\boldsymbol{x}')\cdot \sum_{m=0}^{k} T_m'(\tanh(h_j(\boldsymbol{x}))) \cdot T_m'(\tanh(h_j(\boldsymbol{x}'))) \Bigg].
    \end{array}
\end{equation}
We denote the inner expectation as $D(\boldsymbol{x}, \boldsymbol{x}')$, defined by,
\begin{equation}
D(\boldsymbol{x}, \boldsymbol{x}') := \mathbb{E}_{h, h'}\left[
(1-\tanh^2(h)) \cdot (1-\tanh^2(h')) \cdot \psi(\boldsymbol{x}')\cdot \sum_{m=0}^{k} T_m'(\tanh(h)) \cdot T_m'(\tanh(h'))
\right],
\end{equation}
where $(h, h') \sim \mathcal{N}\left(0, \begin{bmatrix} \sigma^2(\boldsymbol{x}) & \rho(\boldsymbol{x}, \boldsymbol{x}') \\ \rho(\boldsymbol{x}, \boldsymbol{x}') & \sigma^2(\boldsymbol{x}') \end{bmatrix} \right)$. The variance and covariance terms are defined as (see Equations~\eqref{Eq:VarianceH} and~\eqref{Eq:CovH}),
\begin{equation}
\sigma^2(\boldsymbol{x}) := \mathrm{Var}[h_j(\boldsymbol{x})], \quad
\rho(\boldsymbol{x}, \boldsymbol{x}') := \mathrm{Cov}[h_j(\boldsymbol{x}), h_j(\boldsymbol{x}')].
\end{equation}
The derivatives $T_m'(\cdot)$ are evaluated elementwise inside the expectation, as they depend on the random variable $\tanh(h)$, a nonlinear transformation of the Gaussian variable $h$.
Since $h_j(\boldsymbol{x})$ are i.i.d. across neurons, the expectation becomes independent of the index $j$, yielding the simplified form,
\begin{equation}\label{Eq:StepII}
\mathbb{E}[K^{(1)}(\boldsymbol{x}, \boldsymbol{x}')] = N \cdot \sum_{i=1}^d \sum_{n=0}^k T_n(\tilde{x}_i) \cdot T_n(\tilde{x}'_i) \cdot D(\boldsymbol{x}, \boldsymbol{x}').
\end{equation}

Using Equations~\eqref{Eq:StepI} and \eqref{Eq:StepII}, we have, 
\begin{equation}
    \mathbb{E}[\boldsymbol{K}_{\text{ntk}}(\boldsymbol{x}, \boldsymbol{x}')] = N \cdot \left[
\sum_{n=0}^k C_n(\boldsymbol{x}, \boldsymbol{x}') + \sum_{i=1}^d \sum_{n=0}^k T_n(\tilde{x}_i) \cdot T_n(\tilde{x}'_i) \cdot D(\boldsymbol{x},\boldsymbol{x}')
\right],
\end{equation}
where $C_n(\boldsymbol{x},\boldsymbol{x}')$ is correlation of Chebyshev polynomials evaluated at $\tanh(h_j(\boldsymbol{x}))$ and $\tanh(h_j(\boldsymbol{x}'))$, and $D(\boldsymbol{x},\boldsymbol{x}')$ is expectation involving $\psi$ terms and Chebyshev polynomial derivatives.

\newpage
\section{Proof of Theorem \ref{Theorem:NTKConv}}\label{App.Proof2}

    Let $f(\boldsymbol{x}; \boldsymbol{\theta})$ be the output of a cKAN with parameters $\boldsymbol{\theta}$, and let $\boldsymbol{K}_{\text{ntk}}^{(\tau)}(\boldsymbol{x}, \boldsymbol{x}')$ denote the NTK between two inputs $\boldsymbol{x}$ and $\boldsymbol{x}'$ at training time $\tau$. By definition,
    \begin{equation}
        \boldsymbol{K}_{\text{ntk}}^{(\tau)}(\boldsymbol{x}, \boldsymbol{x}') = \left\langle \nabla_{\boldsymbol{\theta}} f(\boldsymbol{x}; \boldsymbol{\theta}(\tau)), \nabla_{\boldsymbol{\theta}} f(\boldsymbol{x}'; \boldsymbol{\theta}(\tau)) \right\rangle.
    \end{equation}
    We now differentiate this quantity with respect to the training time $\tau$ using the product rule,
    \begin{equation}
        \frac{d}{d\tau} \boldsymbol{K}_{\text{ntk}}^{(\tau)}(\boldsymbol{x}, \boldsymbol{x}') = \left\langle \nabla^2_{\boldsymbol{\theta}} f(\boldsymbol{x}; \boldsymbol{\theta}(\tau)) \cdot \dot{\boldsymbol{\theta}}(\tau), \nabla_{\boldsymbol{\theta}} f(\boldsymbol{x}'; \boldsymbol{\theta}(\tau)) \right\rangle + \left\langle \nabla_{\boldsymbol{\theta}} f(\boldsymbol{x}; \boldsymbol{\theta}(\tau)), \nabla^2_{\boldsymbol{\theta}} f(\boldsymbol{x}'; \boldsymbol{\theta}(\tau)) \cdot \dot{\boldsymbol{\theta}}(\tau) \right\rangle.
    \end{equation}
    Taking the absolute value and applying the Cauchy–Schwarz inequality and the uniform bounds $B_1, B_2$, we obtain the bound,
    \begin{equation}
        \left| \frac{d}{d\tau} \boldsymbol{K}_{\text{ntk}}^{(\tau)}(\boldsymbol{x}, \boldsymbol{x}') \right| \leq 2 B_1 B_2 \cdot \left\| \dot{\boldsymbol{\theta}}(\tau) \right\|.
    \end{equation}
    Integrating from $0$ to $\tau$, we get,
    \begin{equation}
        \left| \boldsymbol{K}_{\text{ntk}}^{(\tau)}(\boldsymbol{x}, \boldsymbol{x}') - \boldsymbol{K}_{\text{ntk}}^{(0)}(\boldsymbol{x}, \boldsymbol{x}') \right| \leq 2 B_1 B_2 \cdot \int_0^\tau \left\| \dot{\boldsymbol{\theta}}(s) \right\| ds = 2 B_1 B_2 \cdot \left\| \boldsymbol{\theta}(\tau) - \boldsymbol{\theta}(0) \right\|.
    \end{equation}
    Therefore, the matrix norm satisfies,
    \begin{equation}
        \left\| \boldsymbol{K}_{\text{ntk}}(\tau) - \boldsymbol{K}_{\text{ntk}}(0) \right\| \leq C \cdot \left\| \boldsymbol{\theta}(\tau) - \boldsymbol{\theta}(0) \right\|,
    \end{equation}
    for some constant $C = 2 B_1 B_2$. In the infinite-width limit, parameter drift satisfies,
    \begin{equation}
        \left\| \boldsymbol{\theta}(\tau) - \boldsymbol{\theta}(0) \right\| \to 0 \quad \text{as } N \to \infty,
    \end{equation}
    which implies,
    \begin{equation}
        \left\| \boldsymbol{K}_{\text{ntk}}(\tau) - \boldsymbol{K}_{\text{ntk}}(0) \right\| \to 0 \quad \text{as } N \to \infty.
    \end{equation}

\end{document}